\documentclass[lettersize,journal]{IEEEtran}
\usepackage{amsmath,amsfonts}
\usepackage{algorithmic}
\usepackage{algorithm}
\usepackage{array}
\usepackage[caption=false,font=normalsize,labelfont=sf,textfont=sf]{subfig}
\usepackage{textcomp}
\usepackage{stfloats}
\usepackage{pifont}
\usepackage{url}
\usepackage{xcolor}
\usepackage{footmisc}
\usepackage{xcolor}
\usepackage{verbatim}
\usepackage[T1]{fontenc}
\usepackage{graphicx}
\usepackage{booktabs}
\usepackage{cite}
\usepackage{multirow}
\usepackage{makecell}
\usepackage{hyperref}

\begin{document}

\title{EndoARSS: Adapting Spatially-Aware Foundation Model for Efficient Activity Recognition and Semantic Segmentation in Endoscopic Surgery}

\author{Guankun Wang$^{1,2*}$ \thanks{* Equal contribution.}, Rui Tang$^{3*}$, Mengya Xu$^{1,2*}$, Long Bai$^{1,2}$, Huxin Gao$^{1,2}$, and Hongliang Ren$^{1,2\dagger}$
\thanks{$^\dagger$ Corresponding Author: Hongliang Ren (hlren@ieee.org).}
\thanks{$^{1}$ Department of Electronic Engineering, The Chinese University of Hong Kong, Hong Kong SAR, China.}
\thanks{$^{2}$ Shenzhen Research Institute, The Chinese University of Hong Kong, Shenzhen, China.}
\thanks{$^{3}$ School of Advanced Manufacturing, Fuzhou University, Fuzhou, China.}
}

\markboth{}%
{Shell \MakeLowercase{\textit{et al.}}: }

\maketitle

\begin{abstract} 
Endoscopic surgery is the gold standard for robotic-assisted minimally invasive surgery, offering significant advantages in early disease detection and precise interventions. However, the complexity of surgical scenes, characterized by high variability in different surgical activity scenarios and confused image features between targets and the background, presents challenges for surgical environment understanding. Traditional deep learning models often struggle with cross-activity interference, leading to suboptimal performance in each downstream task. To address this limitation, we explore multi-task learning, which utilizes the interrelated features between tasks to enhance overall task performance. In this paper, we propose EndoARSS, a novel multi-task learning framework specifically designed for endoscopy surgery activity recognition and semantic segmentation. Built upon the DINOv2 foundation model, our approach integrates Low-Rank Adaptation to facilitate efficient fine-tuning while incorporating Task Efficient Shared Low-Rank Adapters (TESLA) to mitigate gradient conflicts across diverse tasks. Additionally, we introduce the Spatially-Aware Multi-Scale Attention that enhances feature representation discrimination by enabling cross-spatial learning of global information within complex surgical environments.In order to evaluate the effectiveness of our framework, we present three novel datasets, MTLESD, MTLEndovis and MTLEndovis-Gen, tailored for endoscopic surgery scenarios with detailed annotations for both activity recognition and semantic segmentation tasks. Extensive experiments demonstrate that EndoARSS achieves remarkable performance across multiple benchmarks, significantly improving both accuracy and robustness in comparison to existing models. These results underscore the potential of EndoARSS to advance AI-driven endoscopic surgical systems, offering valuable insights for enhancing surgical safety and efficiency. Our code and data are available at~\hyperlink{https://github.com/gkw0010/EndoARSS}{https://github.com/gkw0010/EndoARSS}.
\end{abstract}

\begin{IEEEkeywords}
Multi-task learning, Endoscopic surgery, Foundation model, Low-rank adaptation, Surgery understanding.
\end{IEEEkeywords}

\section{Introduction}

Endoscopy has become the gold standard for therapeutic interventions, offering significant improvements in early disease detection and advancing robotic-assisted minimally invasive surgery (RMIS)~\cite{wang2023rethinking}. During endoscopic procedures, the endoscope serves as a critical tool for visualizing surgical instruments and tissues. With the video feed projected onto an external display~\cite{ali2020endoscopy}, endoscopy enables the surgeon to monitor the intraoperative environment. Despite these advancements, endoscopic surgery remains technically demanding with the high risk of bleeding and perforation complications, requiring both highly skilled surgeons and instruments that offer exceptional flexibility and precision~\cite{odagiri2017complications, yamamoto2009endoscopic}. Augmenting the display with intelligently selected key targets in the endoscopic view can further enhance the surgeon's capabilities. To avoid overwhelming the surgical scene with extraneous information, it is crucial to identify and prioritize relevant objects within the scene~\cite{gao2022savanet}. This can enable a comprehensive understanding of the anatomical structures being visualized and manipulated~\cite{allan2020endovis18}. The rapid advancements in deep learning (DL) have significantly enhanced the development of RMIS, particularly through semantic segmentation~\cite{islam2020learning, wang2023rethinking, wang2023domain}. Through leveraging pre-annotated endoscopic surgery datasets, DL models can learn to recognize and infer the positions and interactions between surgical instruments and anatomical structures, thereby providing surgeons with enhanced guidance for navigating complex surgical environments and performing precise interventions~\cite{bai2023surgical, seenivasan2022global}.

\begin{figure*}[!t]
\centering
\includegraphics[width=0.95\textwidth, trim = 20 315 200 0]{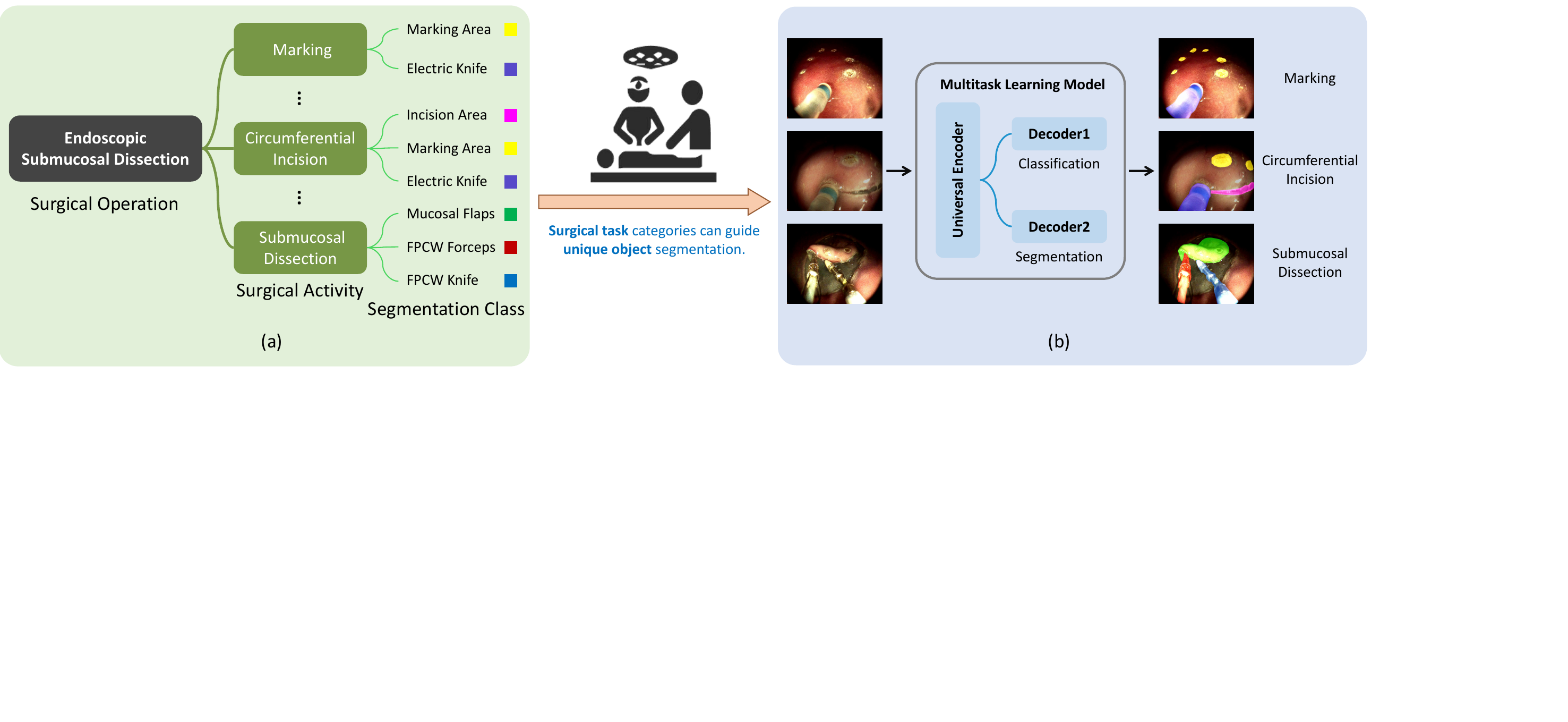}
\caption{Endoscopic submucosal dissection activity information can be utilized to improve performance in cross-activity segmentation. (a) Examples of ESD activity and segmentation category distribution. (b) MTL model for classification and segmentation. Both tasks have the same input data, which uses interaction among multiple task features to boost all tasks' performance.}
\label{fig_intro}
\end{figure*}

\begin{figure}[!t]
\centering
\includegraphics[width=0.48\textwidth]{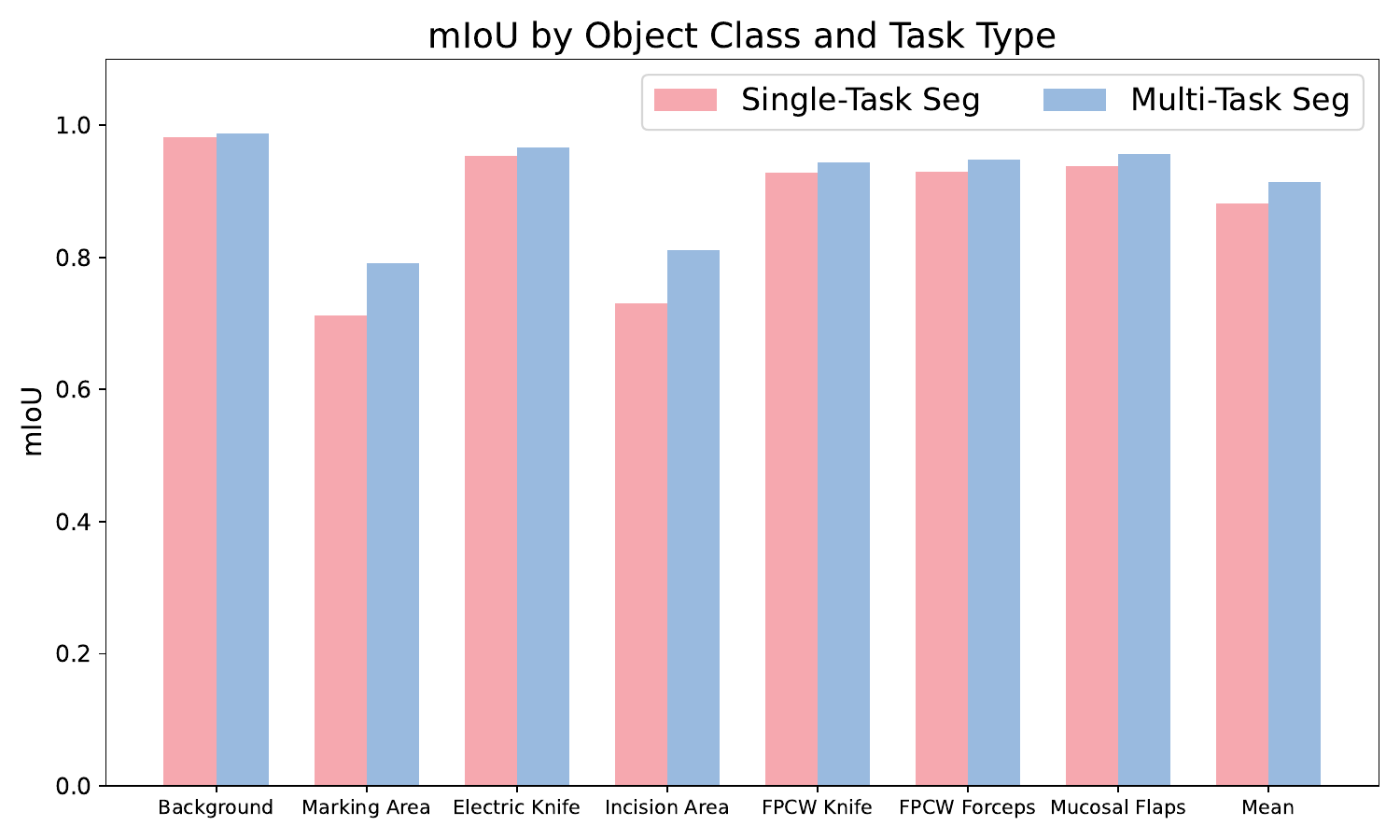}
\caption{Segmentation results of DINOv2~\cite{oquab2023dinov2} and MTL model (DINOv2 serves as the encoder with MMoE~\cite{ma2018modeling} as MTL architecture and STCH~\cite{lin2024smooth} as optimization strategy) on our proposed MTLESD dataset. 
We can observe that the MTL model incorporating the classification task significantly outperforms the model specialized in segmentation.}
\label{fig_introcom}
\end{figure}

However, endoscopic surgery typically involves multiple activities, each distinguished by the utility of specific surgical instruments and interaction with various tissues~\cite{psychogyios2023sar,wang2024copesd}. When a DL model is applied to the entire surgery, it is exposed to all categories of targets present across different activities. Due to the inherent complexity of endoscopic surgical scenes and the visual similarities between different target categories, traditional segmentation approaches are prone to confusion from unrelated regions, leading to ambiguous segmentation outcomes. Figure~\ref{fig_intro}(a) illustrates examples of Endoscopic submucosal dissection (ESD) activities and the distribution of segmentation categories. Consequently, the performance of segmentation models is often limited by cross-activity differences. In actual surgeries, surgeons first identify the ongoing activity before selecting the appropriate instruments and tissues for manipulation. Therefore, the recognition of surgical activities in RMIS has emerged as a hot topic in medical research, enabling surgeons to understand the surgical stages and providing real-time insights into the tasks at hand~\cite{tan2024endoood, bai2024ossar}. These insights could be used to improve cross-activity semantic segmentation. Motivated by this, we investigate the training of a unified model capable of both endoscopic surgical activity recognition and semantic segmentation. Multi-task learning (MTL) is similar to the cognitive processes of the human brain, enabling the handling of multiple tasks, the generalization of acquired knowledge, and learning under limited supervision~\cite{xu2021artificial}. MTL methods exploit the interrelated features between tasks through simultaneously optimizing multiple loss functions within a single model~\cite{haque2021multimix}. By circumventing the need for task-specific feature relearning, MTL promotes the extraction of more generalized representations. This is achieved through the aggregation of shared patterns across tasks, effectively reducing task-specific noise and consequently enhancing overall task performance. Figure~\ref{fig_introcom} compares the performance of a specialized semantic segmentation model with that of an MTL model in cross-activity segmentation, demonstrating significant improvements in segmentation accuracy achieved through the shared feature interaction between classification and segmentation tasks.

Recently, foundation models have garnered considerable attention within the DL community~\cite{wang2023sam, cui2024surgical}. These models, characterized by their extensive parameterization, exhibit remarkable capacity for long-term memory retention across expansive training datasets, achieving state-of-the-art (SOTA) performance across a variety of downstream tasks involving visual, textual, and multimodal inputs. Considering advanced capabilities for the extraction of unified visual features, we select DINOv2~\cite{oquab2023dinov2} as the foundational backbone for developing an MTL perception system tailored for endoscopic surgery. However, foundation models face a severe degradation in predictive performance when applied them to specific domains~\cite{wang2023sam}. Developing a specialized medical foundation model from scratch presents substantial challenges since the available data and computational resources are limited. To address these limitations, we introduce the adaptation strategy utilizing Low-Rank Adaptation (LoRA)~\cite{hu2021LoRA}.
Instead of the entire model fine-tuning, we freeze the foundation model while integrating trainable LoRA layers. 
Besides, data heterogeneity, an intrinsic characteristic of MTL, also poses a significant challenge as it can induce training conflicts between tasks. In this case, we propose our Endoscopy Surgery Activity Recognition and Semantic Segmentation (EndoARSS) framework with Task Efficient Shared Low-rank Adapter (TESLA) which integrates task-specific LoRA into the backbone to facilitate parameter isolation between different tasks and alleviate training conflicts. Furthermore, EndoARSS employs Spatially-Aware Multi-Scale Attention (SMA) which is designed to generate more discriminative feature representations in complex endoscopic surgery environments. Overall, our key contributions can be summarized as follows:
\begin{itemize}
    \item[--] We propose the EndoARSS, a multi-task learning framework designed for activity recognition and semantic segmentation in endoscopic surgery.  Built upon the DINOv2 foundation model, EndoARSS effectively uses multi-task features and their interactions to achieve enhanced performance on both tasks.
    
    \item[--] We design the Task Efficient Shared Low-rank Adapters (TESLA), which integrates two task-specific LoRA adjacent to the foundation model to facilitate parameter isolation between two tasks. This method efficiently reduces gradient conflicts during training by employing task-specific low-rank parameters, enabling the model to handle heterogeneous data effectively.

    \item[--] To address the challenges of feature ambiguity in complex surgical environments, we introduce the Spatially-Aware Multi-Scale Attention (SMA) mechanism. SMA encodes both local and global information within a tensor, allowing for cross-spatial learning of global information. This mechanism helps the model generate more discriminative feature representations to mitigate feature ambiguity in endoscopic surgery.
    
    \item[--] To evaluate the effectiveness and generalization capability of EndoARSS, we propose three multi-task datasets: MTLESD, MTLEndovis and MTLEndovis-Gen. They enable extensive evaluation of both surgical activity recognition and semantic segmentation tasks in various endoscopic surgical scenarios.
    
    \item[--] Extensive experiments are conducted on our proposed datasets, demonstrating that EndoARSS can achieve superior multi-task performance, underscoring its capability to provide a robust, high-level understanding of complex endoscopic surgery scenarios.
    
\end{itemize}

\section{Related Work}
\subsection{Multi-Task Learning}

MTL approaches can generally be classified into two classes: hard parameter sharing (HPS)~\cite{caruana1993multitask} and soft parameter sharing (SPS)~\cite{duong2015low}. HPS techniques~\cite{long2017learning,lu2017fully} consolidate feature extractors across all tasks and may employ multi-objective optimization (MOO)~\cite{kendall2018multi, chen2018gradnorm, sener2018multi} to fine-tune the weights of various losses. While these methods maintain scale invariance for numerous tasks, they tend to favor those with stronger signals. In contrast, SPS approaches~\cite{misra2016cross, liu2019end, wallingford2022task} use interactions among multiple feature extractors to boost task performance, though they significantly increase parameter size as the number of tasks grows. More recently, Mixture of Experts (MoE) frameworks~\cite{shazeer2017outrageously, fan2022m3vit, aoki2022heterogeneous} have implemented gate functions to allocate distinct parameter sets to individual tasks, achieving notable results. Nevertheless, they share the same challenges as SPS, including a high parameter count and scaling issues, along with a lack of interpretability regarding the role of experts within the MoE framework. 


Based on the advantage of shared feature learning, MTL has demonstrated remarkable performance in computer vision, facilitating tasks like object detection, semantic segmentation, and depth estimation, etc. Notable works include Cross-Stitch Networks~\cite{misra2016cross} that introduced learnable feature fusion to enable adaptive sharing across tasks. UniT~\cite{hu2021unit} unified diverse vision tasks (detection, segmentation) via a single transformer, demonstrating cross-task knowledge transfer. TaskPrompter~\cite{ye2022taskprompter} used learnable prompts to generalize MTL to unseen tasks, while OFTNet~\cite{wu2022orthogonal}proposed orthogonal feature transformations for disentangled task representations. Regarding MTL in medical image analysis, most works involve the segmentation task~\cite{zhu2020lymph}, while others focus on tasks like classification~\cite{grimwood2020assisted,ramesh2021multi}, registration~\cite{du2020multi}, landmark detection~\cite{xu2018less}, survival prediction~\cite{liu2020prediction, yao2021deepprognosis}, and report generation~\cite{seenivasan2023task}. For instance, a parallel-based MTL model has been designed to achieve simultaneous segmentation of gastric tumors and classification of lymph nodes. This model includes two subnets specifically designed for different tasks, which use common features across distinct tasks and incorporate a task-specific attention-driven learning module~\cite{zhang20213d}. An MTL model~\cite{song2020end} simultaneously segments and classifies skin lesions. By utilizing shared representations, existing works demonstrated improved performance compared to single-task architectures, highlighting the benefits of knowledge sharing across related tasks. However, they often suffer from inefficient parameters and inadequate handling of gradient conflicts in complex task settings. Therefore, we aim to develop a unified framework applying low-rank adaptation of large foundation models to improve parameter efficiency and mitigate gradient interference via task-specific adapters.


\subsection{Low-Rank Adaptation Empowers Foundation Model}
Foundation models have emerged as a transformative approach in various fields, leveraging vast amounts of data to develop generalized representations that can be adjusted for different downstream tasks. These models offer a robust foundation for various applications and excel in identifying complicated patterns and connections within the data. Notable examples include models like BERT~\cite{kenton2019bert}, and GPT~\cite{radford2019language}, demonstrating significant performance improvements compared to traditional methods. SAM~\cite{kirillov2023segment} and SAM2~\cite{ravi2024sam} use large-scale datasets and advanced self-supervised learning techniques to learn generalized segmentations. DINOv2~\cite{oquab2023dinov2} uses self-supervised learning to enhance representation quality.

However, training foundation models requires high resource demands and these models have billions of parameters, necessitating large memory and storage capacities. Maintaining stability during training to prevent issues like model collapse or gradient vanishing is essential. These challenges highlight the need for research in training techniques to make foundation models more efficient and accessible. Low-Rank Adaptation (LoRA)~\cite{hu2021LoRA} is a noteworthy technique designed to enhance the efficiency of fine-tuning foundation models. LoRA enables practitioners to adapt large pre-trained models with fewer resources while maintaining performance by introducing low-rank parameter updates. This method effectively reduces trainable parameters amount, thereby reducing computational costs and data memory demand. Recent studies~\cite{wang2023sam, wu2024fedfmsl, wang2024surgical, chen2024robust} have illustrated that LoRA can achieve competitive results across various tasks, making it an appealing choice aiming to use the power of foundation models without incurring prohibitive expenses. RoLoRA~\cite{chen2024robust} explores the integration of LoRA with federated learning of Foundation Models. LoRA is also utilized to apply a foundation model pre-trained on satellite imagery~\cite{selvam2024rapid}. Explained Variance Adaptation (EVA)~\cite{paischer2024one} enhances LoRA by initializing new weights data-driven using singular value decomposition. The synergy between foundation models and adaptation techniques like LoRA highlights a significant trend in DL research. By enabling efficient fine-tuning and enhancing model versatility, these methodologies open up new possibilities for developing intelligent systems that can tackle various challenges across various sectors.


\section{Methodology}
In this work, we propose the EndoARSS framework (shown in Figure~\ref{fig_main}), which is built upon the DINOv2~\cite{oquab2023dinov2}. Given an endoscopic surgery image $x \in \mathbb{R}^{H \times W \times C}$, EndoARSS aims to simultaneously predict the segmentation mask $m \in \mathbb{R}^{H \times W}$ and surgical activities. DINOv2 functions as the image encoder, where images first undergo Spatially-Aware Multi-Scale Attention (SMA) to produce discriminative features. The processed image features are subsequently divided into patches, followed by flattening and linear projection. Then, we add a positional embedding and a learnable class token, which facilitates the aggregation of global image information. The resulting image embeddings are propagated through a sequence of Transformer layers, which generate updated token representations. Notably, the parameters of the DINOv2 are frozen during training. LoRA layers are integrated into each Transformer layer to capture learnable knowledge. As outlined in the previous section, these side LoRA layers compress the Transformer features into a low-rank space, followed by re-projection to align with the output channels of the Transformer layers. Notably, each LoRA layer operates independently within its corresponding Transformer layer, without weight sharing. Furthermore, we propose Task Efficient Shared Low-rank Adapters (TESLA) to address data heterogeneity while training the unified model for two tasks. Inspired by the decoder of DeepLab~\cite{chen2017deeplab}, we adapt decoder heads to predict segmentation masks and surgical activities, respectively.

\subsection{Preliminaries}

\subsubsection{DINOv2}
The utility of pre-trained representations without task-specific adaptation has been thoroughly validated in the field of Natural Language Processing~\cite{raffel2020exploring}. Notably, features extracted from these models can effectively serve downstream tasks without necessitating fine-tuning, achieving superior performance to that of tailored models. In the domain of Computer Vision, \cite{oquab2023dinov2} introduced a foundational model named as DINOv2, which produces visual features at both image and pixel levels, independent of task constraints. DINOv2 is an automated framework for creating a large, curated dataset designed for image data, alongside an unsupervised learning approach for robust visual feature extraction. As a Vision Transformer (ViT) model comprising 1 billion parameters, DINOv2 is initially trained via a discriminative self-supervised learning approach. Subsequently, this large-scale model is distilled into several smaller models. These distilled models demonstrated superior performance compared to the best existing general-purpose features across a wide array of benchmarks, both at the image and pixel scales.

\subsubsection{Multi-Task Parameters}
For each transformer layer within the backbone, we define its weight as $\mathbf{W}_\mathbf{0} \in \mathbb{R}^{C^{\text{out}} \times C^{\text{in}} \times k \times k}$, where $C^{\text{out}}$ and $C^{\text{in}}$ refer to the output and input channels, respectively, while $k$ represents the kernel size. Let $x \in \mathbb{R}^{b \times C^{\text{in}} \times H \times W}$ and $h \in \mathbb{R}^{b \times C^{\text{out}} \times H \times W}$ represent the input token embedding and their corresponding outputs, where $b$ is the batch size and $H$ and $W$ correspond to the height and width of the feature maps. $x_t$ and $h_t$ are the input and output features of the $t$-th task within the current mini-batch.

\begin{figure*}[!t]
\centering
\includegraphics[width=0.9\textwidth, trim = 0 0 300 0]{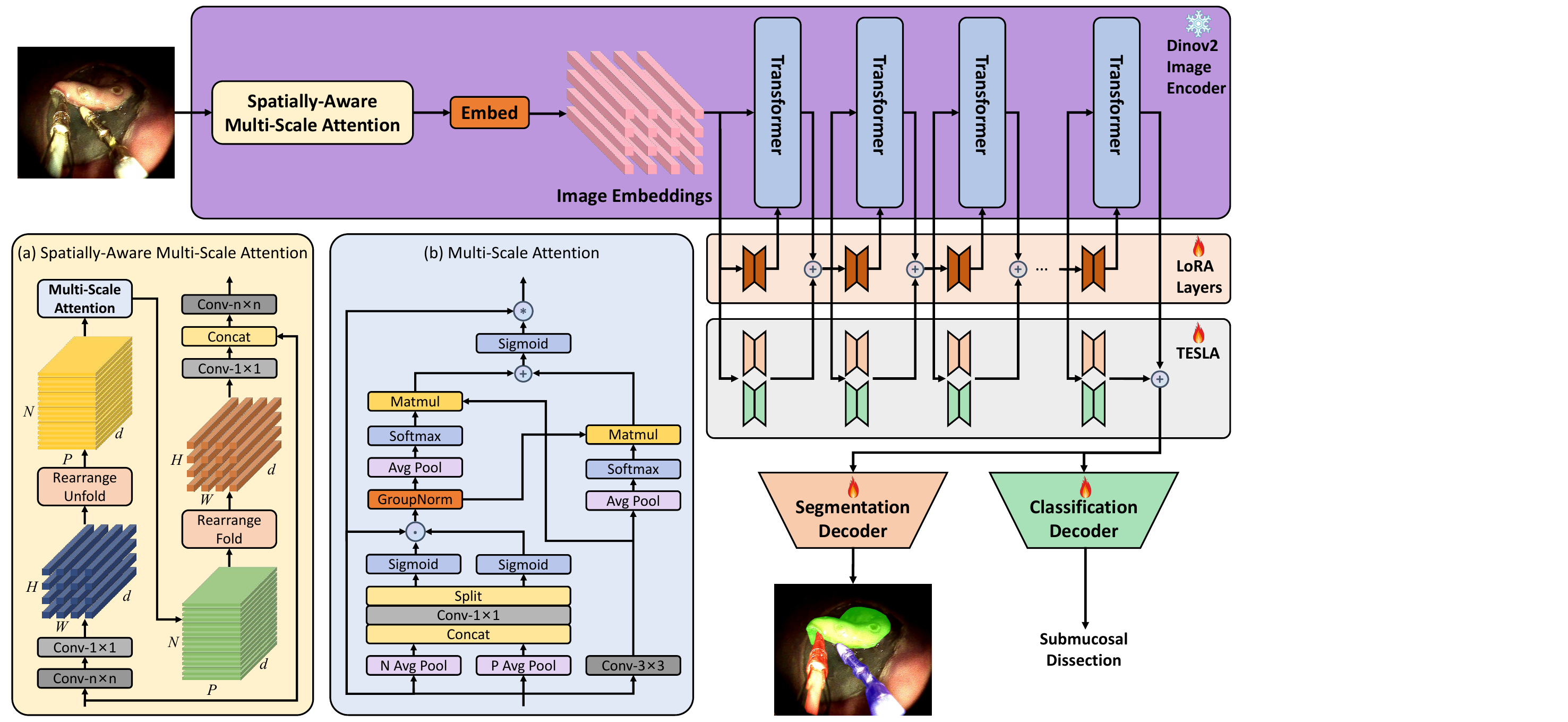}
\caption{Overview of our proposed EndoARSS framework. EndoARSS comprises spatial-aware multi-scale attention, freezed DINOv2 image encoder, trainable LoRA layers, task-efficient shared low-rank adapters and two parallel prediction heads for activity recognition and semantic segmentation. $\oplus$ and $\odot$ represent the summation operation and element-wise dot product, respectively.}
\label{fig_main}
\end{figure*}

\subsection{Spatial-aware Multi-scale Attention}
In complex surgical environments, the high similarity between tissue and intricate surgical backgrounds poses significant challenges. Additionally, various surgical instruments often exhibit comparable mechanical designs. These factors result in a lack of clear distinction in image features among different objects. Such interference hampers the ability of models to accurately differentiate between target and non-target regions, as well as among distinct target areas, ultimately impacting the precision of surgical scene understanding. To mitigate this issue, we introduce Spatially-Aware Multi-Scale Attention (SMA), which facilitates cross-spatial learning of global representations to produce more discriminative features. Figure~\ref{fig_main} (a) provides the details of SMA.

\subsubsection{Unfolding for Learning Global Representations}
SMA first encodes both local and global information within the input tensor while minimizing parameter usage. Specifically, the input tensor $x_o \in \mathbb{R}^{H \times W \times C}$ is successively processed through a $n \times n$ convolutional layer and a $1 \times 1$ convolutional layer to produce $x_l \in \mathbb{R}^{H \times W \times d}$. The former convolution captures local spatial features, and the latter layer projects the tensor into a high-dimensional space through linear combinations of input channels. Among self-attention mechanisms, vision transformers (ViTs) equipped with multi-head self-attention effectively handle non-local and long-range dependencies, boasting the receptive field of $H \times W$. However, ViTs are computationally intensive and exhibit limited optimizability since they lack spatial inductive bias. To learn global representations with spatial inductive bias while utilizing fewer parameters, we incorporate standard convolution operations: unfolding, global processing, and folding~\cite{mehta2021mobilevit}. Processed tensor $x_l$ is first unfolded into $N$ non-overlapping flattened patches, denoted as $x_u \in \mathbb{R}^{P \times N \times d}$. $P$  equal to $wh$. $N$ is calculated as $\frac{HW}{P}$ representing the number of patches, with patch height $h \leq n$ and width $w \leq n$.

\subsubsection{Multi-Scale Attention Module}
For each $p \in\{1, \cdots, P\}$ in non-overlapping flattened patches $x_u$, inter-patch relationships are processed through the multi-scale attention module~\cite{ouyang2023efficient} to yield $x_d \in \mathbb{R}^{P \times N \times d}$ as follows:
\begin{equation}
x_d\left(p\right)=\mathcal{A}\left(x_u\left(p\right)\right), 1 \leq p \leq P
\end{equation}
As illustrated in Figure~\ref{fig_main} (b), we use the feature grouping mechanism alongside the multi-scale structures of multi-scale attention to effectively model short and long-range dependencies for discriminative feature generation. For the input $x_u \in \mathbb{R}^{P \times N \times d}$, we implement three parallel pathways to extract attention weight descriptors: two pathways operate as $1 \times 1$ branches, and the third employs a $3 \times 3$ kernel. For efficiently capturing dependencies across channels, two 1D global average pooling are applied, followed by two spatial directions within the $1 \times 1$ branches. The multi-scale feature representations are captured by $3 \times 3$ branch with a $3 \times 3$ kernel. Subsequently, the pooled features are concatenated and processed through a $1 \times 1$ convolution, maintaining dimensionality. The output is factorized into two vectors, each of which is modulated by a non-linear Sigmoid function to approximate a 2D binomial distribution leveraging linear convolutions. To perform cross-channel interactions between two $1 \times 1$ branches, we perform a simple multiplication of each group's channel-wise attention map. Conversely, the third branch facilitates cross-channel interactions through a $3 \times 3$ convolution, expanding the feature space. The multi-scale attention module encodes inter-channel information to modulate channel importance and preserve precise spatial structure. Furthermore, it enhances feature aggregation by providing cross-spatial information across various spatial dimensions. These branches' outputs undergo 2D global average pooling, after which a Softmax function is applied to fit 2D Gaussian maps. The final output for each group is computed by aggregating the spatial attention weights derived from the two pathways, followed by the Sigmoid function. Such an operation effectively captures pixel-level pairwise relationships while emphasizing global context. Since the resulting feature map retains the original dimensions of $x_u$, it is efficient and effective to integrate it into DINOv2.

As mentioned above, the integration of multi-scale contextual information empowers the module to deliver enhanced pixel-level attention in high-level feature maps, thereby enhancing the discriminative representation for different target regions. Additionally, the convolutional kernel parallelization proves to be a potent approach for addressing short and long-range dependencies through cross-spatial learning methods.

\subsubsection{Folding Back Encoded Representations into Spatial Dimensions}
Since the multi-scale attention module preserves both the spatial order of pixels and the patch order within each patch, the downsampled representation $x_d$ can be folded to yield $x_f \in \mathbb{R}^{H \times W \times d}$. $x_f$ is subsequently projected into the lower-dimensional space via $1 \times 1$ convolution layer, after which it is concatenated with the original input tensor $x$. To integrate the concatenated features, an additional $n \times n$ convolution layer is applied. Given that $x_d(p)$ encodes global information across $P$ patches, each pixel in $x_d$ can encode information from all pixels in $x$. Thus, the overall receptive field of SMA is $H \times W$.

\subsection{Low-rank Adapter Layers for Model Fine-tuning}

In contrast to the entire model fine-tuning, we freeze the foundation model while incorporating trainable LoRA layers, which can significantly reduce the computational resources and memory during training, facilitating easier model deployment. The LoRA within EndoARSS is illustrated in Figure~\ref{fig_main}. Specifically, we compute the weight updates of a weight matrix $\mathbf{W}_\mathbf{0}$ as $\Delta \mathbf{W}$ in LoRA. $\Delta \mathbf{W}$ is approximated through the matrix multiplication of two smaller matrices $BA$:
\begin{equation}
h=\mathbf{W}_\mathbf{0} x+\Delta \mathbf{W} x=\mathbf{W}_\mathbf{0} x+B A x = \mathbf{W}^{\prime} x
\end{equation}
where $x$ is the token embedding. Following the approach of~\cite{zhang2023customized}, we apply low-rank approximation exclusively to the $q$ and $v$ projection layers to minimize their impact on attention scores. Utilizing above LoRA foundational operation, the computation of the projection layers for $q$, $k$, and $v$ in a multi-head self-attention block is articulated as follows:

\begin{equation}
\begin{array}{l}
Q=\hat{W}_{q} x=W_{q} x+B_{q} A_{q} x, \\
K=W_{k} x, \\
V=\hat{W}_{v} x=W_{v} x+B_{v} A_{v} x 
\end{array}
\end{equation}

The matrices $W_q$, $W_k$, and $W_v$ denote the frozen projection layers corresponding to the queries ($q$), keys ($k$), and values ($v$), respectively. The trainable LoRA layers are represented by $A_q$, $B_q$, $A_v$, and $B_v$. The self-attention mechanism retains its original formulation, and output tokens $C_{out}$ are computed as follows:

\begin{equation}
\operatorname{Att}(Q, K, V)=\operatorname{Softmax}\left(\frac{Q K^{T}}{\sqrt{C_{\text {out }}}}+B\right) V
\end{equation}

\begin{figure}[!t]
\centering
\includegraphics[width=0.45\textwidth]{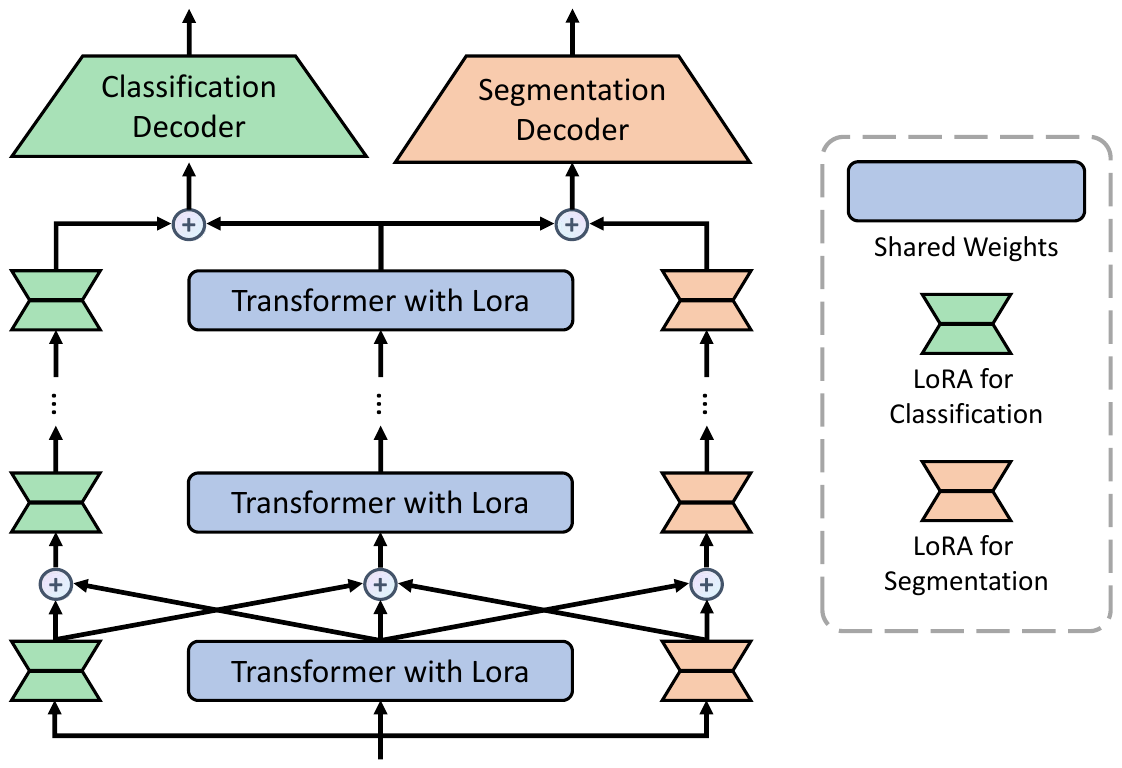}
\caption{The overview of TESLA. Blue rectangles represent shared weights and orange and green hexagons represent task-specific low-rank adapters. $\oplus$ represents the weighted sum of TESLA based on the output of the task-specific low-rank adapters.}
\label{fig_tesla}
\end{figure}

\subsection{Task Efficient Shared Low-rank Adapters}

Training a unified model capable of handling diverse tasks confronts significant challenges due to the presence of data heterogeneity. The discrepancies arising from heterogeneous data result in formidable conflicts during the training phase, impairing the full exploitation of the dataset scale advantages and substantially degrading model's performance~\cite{aoki2022heterogeneous}. Tackling this issue necessitates a carefully crafted approach that fosters a harmonious coalescence of heterogeneous data training conflicts. In this spirit, we utilize the task-specific low-rank adapters that are attached to the shared backbone of the model~\cite{zhou2024exploring}. Connecting multiple low-rank adapters in parallel to the weights of the DINOv2, we propose Task Efficient Shared Low-rank Adapters (TESLA) since it is characterized by controllable parameters and shared knowledge with backbone. The layout of TESLA is elaborated within our EndoARSS, as depicted in Figure~\ref{fig_main}.

By using task-specific low-rank adapters, TESLA combines different training parameters for different tasks. The isolation of these parameters subdues gradient conflicts emerging from different tasks, culminating in an effective mitigation strategy for the heterogeneous training conflict. Significantly, these low-rank adapters drastically curtail the number of introduced parameters, enabling large-scale heterogeneous training. This process, in concatenation with the fusion of the backbone network's parameters, assures that the resulting architecture retains sufficient representation capacity.

For each low-rank adapter, there are two factors: $\mathbf{A}_{\mathbf{i}} \in \mathbb{R}^{r k \times C^{\text {in }} k}$ and $\mathbf{B}_{\mathbf{i}} \in \mathbb{R}^{C^{\text {out }} k \times r k}$, where $i \in\{1, \ldots, E\}$ and $r$ represents the rank. $E$ is the number of adapters. Then, the transformer operation can be transformed into:
\begin{equation}
    h_{t}=\left(\mathbf{W}^{\prime}+\sum_{i=1}^{E} \alpha_{i} \mathbf{B}_{\mathbf{i}} \mathbf{A}_{\mathbf{i}}\right) x_{t}
    \label{eq_LoRA}
\end{equation}

where $\alpha_{i}$ is the weight of the $i$-th low-rank adapter. Equation~\ref{eq_LoRA} represents the foundational form of TESLA, which employs a weighted combination of $E$ low-rank adapters alongside a shared weight matrix $\mathbf{W}^{\prime}$ to allocate distinct parameter sets for different tasks. This parameter isolation helps mitigate heterogeneous conflicts by separating the gradients of various tasks. Gradient separation is intuitively categorized into explicit and implicit types~\cite{zhou2024exploring}, and we use the first type for complex target perception in endoscopic surgery contexts.

TESLA uses task identifiers to explicitly distinguish the gradients of heterogeneous data, thereby reducing training conflicts. Figure~\ref{fig_tesla} illustrates the details of the TESLA. We assign specific low-rank adapters to each task, ensuring that each adapter computes gradients solely for its corresponding task, facilitating explicit gradient separation. Since the shared weight matrix $\mathbf{W}^{\prime}$ is involved in the computations across all tasks, we adopt the idea of group convolution to optimize the calculation process for transformer operations, minimizing redundant computations and conserving memory resources. Specifically, for each transformer layer that incorporates TESLA, the output feature $h$ can be computed by

\begin{equation}
    h=\left(\mathbf{W}^{\prime}+\mathbf{B}_{\mathbf{1}} \mathbf{A}_{\mathbf{1}}\right) x_{1} \cup \left(\mathbf{W}^{\prime}+\mathbf{B}_{\mathbf{2}} \mathbf{A}_{\mathbf{2}}\right) x_{2} \\
    =\mathbf{W}^{\prime\prime} x
    \label{eq_tesla}
\end{equation}

where $\cup$ represents the concatenation operation. Subsequently, we reshape the integrated weight $\mathbf{W}^{\prime\prime} \in \mathbb{R}^{b \times C^{\text {out }}} \times C^{\text {in }} \times k \times k \text { to } \mathbb{R}^{b C^{\text {out }} \times C^{\text {in }} \times k \times k}$, $r$ to $\mathbb{R}^{1 \times b C^{\text {in }} \times H \times W}$, and define the group number to $b$.  


\section{Experiments}
\subsection{Datasets}
\label{sec:datasets}

To effectively evaluate our methods, we propose three multi-task datasets for activity recognition and semantic segmentation based on endoscopic surgery scenarios: Multi-Task Learning for Endoscopic Submucosal Dissection (MTLESD) and Multi-Task Learning for Endovis (MTLEndovis) dataset. The details and dataset distribution are shown in Table.~\ref{tab:distribution}.

\noindent \textbf{MTLESD} is a novel dataset tailored to endoscopic submucosal dissection (ESD). Leveraging DREAMS system~\cite{gao2023transendoscopic,yang2023novel}, we obtain ESD procedure videos performed on in-vivo porcine models. The animal study was approved by the Institutional Ethics Committee on Animal Experiments (Approval No. DWLL-2021-021). These videos are recorded using a dual-channel flexible endoscope with the resolution of $1920 \times 1080$. The final resolution of the endoscopic images is $1306 \times 1009$ after cropping. Expert endoscopists from Qilu Hospital annotated the ESD videos, focusing on six key surgical activities: marking, injection, circumferential incision, subsidized injection, installation and debugging, and submucosal dissection. From these, we select three high-interaction activities—marking, injection, and submucosal dissection—for activity recognition task. Six corresponding segmentation categories related to these selected activities are illustrated in Figure~\ref{fig_intro}. In total, we extracted and annotated 2,020 images. For each activity, the dataset is partitioned into training and test sets in a 4:1 ratio.

\noindent \textbf{MTLEndovis} is derived from the publicly available dataset, EndoVis-18~\cite{allan2020endovis18}, which provides both images and corresponding segmentation masks of the porcine kidney surgery. The activity annotations are sourced from EndoVis-18-VQLA-Extend~\cite{bai2025surgical}. To obtain the surgical activities, we extracted and analyzed the actions of surgical instruments in each frame. This analysis resulted in the identification of 12 distinct surgical activities, namely: cauterization, clipping, cutting, grasping, idle, looping, manipulation, retraction, stapling, suction, suturing, and ultrasound sensing. Given that the images in the EndoVis-18-VQLA-Extend dataset are also collected from EndoVis-18, we use the corresponding segmentation annotations from EndoVis-18, which include 7 instrument categories: Bipolar Forceps, Large Needle Driver, Prograsp Forceps, Ultrasound Probe, Suction Instrument, Monopolar Curved Scissors, and Clip Applier. Ultimately, \textbf{MTLEndovis} comprises a total of 1,996 images, which is the same as EndoVis-18-VQLA-Extend. The training and test sets are randomly partitioned in the same ratio as MTLESD dataset.

\noindent \textbf{MTLEndovis-Gen} is collected from the MICCAI EndoVis Challenge 2017~\cite{allan2019endovis17}, with activity annotations provided by EndoVis-17-VQLA-Extend~\cite{bai2025surgical} and segmentation labels sourced from the original challenge. This dataset is employed to evaluate the generalization capabilities of our model under domain shift conditions. Specifically, we \textit{train} the model exclusively on the MTLEndovis training set and subsequently \textit{evaluate} its performance on MTLEndovis-Gen. Comprising 97 frames, the dataset encompasses 6 surgical activities and 5 segmentation classes, all of which are present in the MTLEndovis dataset.

\begin{table}[]
\caption{Details of our proposed MTL-ESD and MTL-Endovis18 datasets. 'C Number' and 'S Number' indicate the number of categories for classification and segmentation tasks.}
\label{tab:distribution}
\resizebox{0.48\textwidth}{!}{
\renewcommand{\arraystretch}{1}
\begin{tabular}{c|c|c|c|c|c}
 \hline
Dataset       & C Class & S Class & Train & Test  & Total  \\ \hline
MTLESD       & 3       & 6       & 1618  & 402   & 2020   \\ \hline
MTLEndovis & 12      & 7       & 1601  & 395   & 1996   \\ \hline
MTLEndovis-Gen & 6      & 5       & 0  & 97   & 97   \\ \hline
\end{tabular}}
\end{table}

\subsection{Implementation Details}
Multi-task learning models typically contain three components: the backbone, architecture, and optimization strategy~\cite{lin2022libmtl}. In our study, we evaluate a combination of three backbone networks, three optimization algorithms, and four architectural paradigms. Specifically, we consider ResNet101~\cite{he2016deep}, DINOv2~\cite{oquab2023dinov2}, and the EndoARSS Backbone (DINOv2 with SMA) as backbones. For optimization, we utilize MMoE~\cite{ma2018modeling}, DSelect\_k~\cite{hazimeh2021dselect}, and our proposed TESLA. The architecture include MoCo~\cite{fernando2023mitigating}, Aligned-MTL~\cite{senushkin2023independent}, STCH~\cite{lin2024smooth}, and DB-MTL~\cite{lin2023dual}. We combine these components and conduct an in-depth comparative study of our proposed framework with the aforementioned SOTA approaches.

We employ separately customized adaptations of Deeplab-V3~\cite{chen2017rethinking} as decoders for activity recognition and semantic segmentation tasks. For the task of activity recognition, we utilize evaluation metrics such as classification accuracy (AC) and F-score. In the context of segmentation, the performance is quantified through the mean Intersection over Union (mIoU), Dice Similarity (DS), Average Hausdorff Distance (HD), and Structural Similarity Index Measure (SSIM). All implementations utilize the PyTorch framework in Python on NVIDIA RTX 4090 GPU. For the MTLESD dataset, the learning rate is set to $1 \times 10^{-4}$, whereas for the MTLEndovis dataset, it is set to $1 \times 10^{-3}$. For both datasets, training is performed with the Adam optimizer with a batch size of 16 over 200 epochs.

\subsection{Results}
\subsubsection{Evaluation on MTLESD}
\begin{figure*}[!t]
\centering
\includegraphics[width=\textwidth]{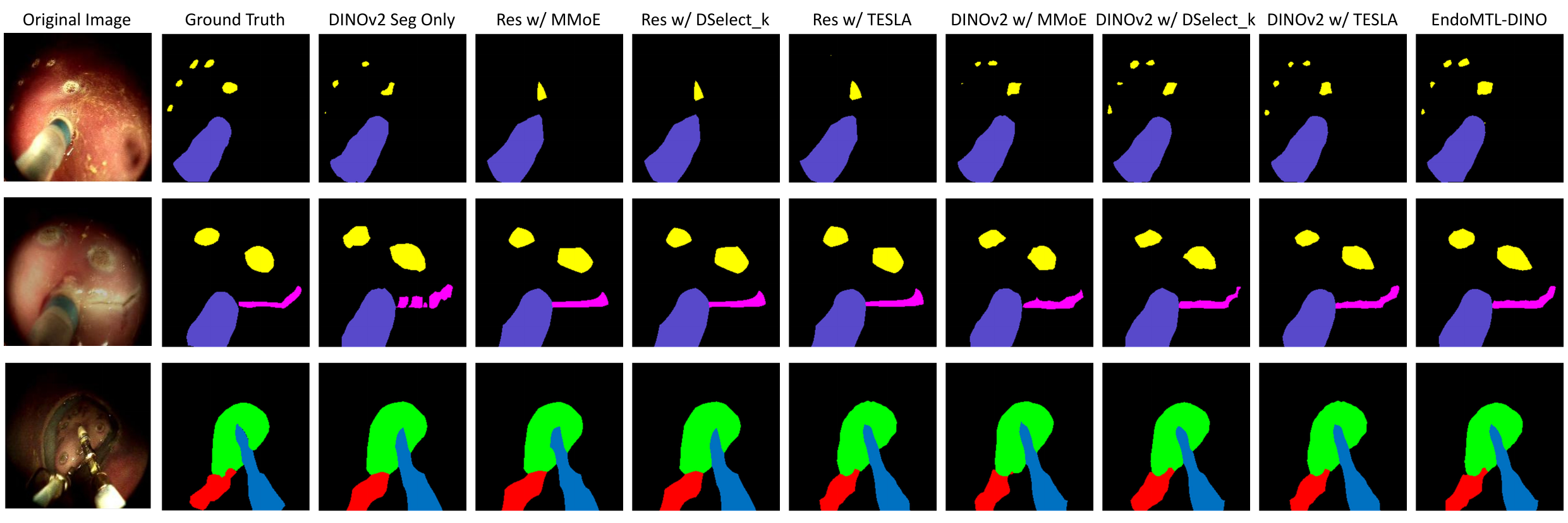}
\caption{Segmentation visualization on the MTLESD dataset. All MTL configurations incorporate DB-MTL as an optimization strategy due to its superior performance on this dataset. For a visual reference regarding the categories associated with distinct colors, please refer to Figure~\ref{fig_intro} (a).}
\label{fig_visual}
\end{figure*}

We first evaluated our proposed EndoARSS framework on the MTLESD dataset. Table~\ref{tab:esd} presents the quantitative experimental results for various MTL combinations, and Figure~\ref{fig_visual} shows the visualization of the segmentation with DB-MTL optimization strategy. For the surgery activity recognition task, all MTL models perform well since there are only three kinds of activities and the scene differences for each activity are huge. However, compared with specialized semantic segmentation, the segmentation performance of DINOv2 has improved significantly after the introduction of the multi-task framework, which proves the effectiveness of shared feature interaction between tasks. Besides, it is notable that models utilizing the DINOv2 backbone consistently outperform those with ResNet101 across all metrics. This performance boost highlights the superior feature extraction capabilities of DINOv2, particularly for complex surgical environments. Moreover, the introduction of the TESLA architecture further enhances performance, particularly when combined with DINOv2. TESLA mitigates gradient conflicts and facilitates parameter isolation across tasks, yielding a maximum mIoU improvement of 0.96\% and a DS increase of 0.62\% in the DB-MTL strategy. This demonstrates that TESLA's efficient parameter sharing is crucial for multi-task learning. 
Notably, the EndoARSS backbone (DINOv2 with SMA) achieves an overwhelming advantage, surpassing all other configurations. It reaches the best mIoU of 92.16\% and DS of 95.79\%, and SSIM of 95.67\% and also lowers the HD to 13.25, outperforming other SOTA methods by significant margins. Compared to the DINOv2 backbone, EndoARSS with SMA can better handle feature ambiguity, contributing to enhanced segmentation performance.
\begin{table*}
\caption{Multi-task learning results on the MTLESD dataset. We combine and compare the selected components one by one with our EndoARSS. Since DB-MTL is the best-performing optimization strategy in the MTLESD dataset, we only provide the results of the EndoARSS backbone (DINOv2 with SMA) based on DB-MTL for comparison with other SOTA combinations.}
\label{tab:esd}
\centering
\resizebox{\textwidth}{!}{
\renewcommand{\arraystretch}{1}
\begin{tabular}{c|cc|cccccc}
\toprule[1pt]
\multirow{2}{*}{Backbone} &
  \multicolumn{2}{c|}{Multi-Task Algorithms} &
  \multicolumn{2}{c|}{Activity Recognition} &
  \multicolumn{4}{c}{Semantic Segmentation} 
 \\ \cline{2-9} 
 &
 \multicolumn{1}{c|}{Architectures} &
 \multicolumn{1}{c|}{\makecell[c]{Optimization \\ Strategies}} &
  AC(\%) &
  \multicolumn{1}{c|}{F-score(\%)} &
  mIoU(\%) &
  DS(\%) &
  HD &
  SSIM(\%) 
   \\ \hline

\multirow{12}{*}{ResNet101} &
  \multicolumn{1}{c|}{\multirow{4}{*}{MMoE}} &
  \multicolumn{1}{c|}{MoCo} &
  99.99 &
  \multicolumn{1}{c|}{99.95} &
  82.08 &
  89.63 &
  18.48 &
  91.19 \\ 
 &
  \multicolumn{1}{c|}{} &
  \multicolumn{1}{c|}{Aligned-MTL} &
  99.99 &
  \multicolumn{1}{c|}{99.95} &
  80.57 &
  88.6 &
  18.83 &
  90.73 \\ 
 &
  \multicolumn{1}{c|}{} &
  \multicolumn{1}{c|}{STCH} &
  99.99 &
  \multicolumn{1}{c|}{99.95} &
  83.37                        & 90.46                        & 18.29                        & \textbf{91.71} \\ 
 &
  \multicolumn{1}{c|}{} &
  \multicolumn{1}{c|}{DB-MTL} &
  \textbf{99.99} &
  \multicolumn{1}{c|}{\textbf{99.95}} &
\textbf{83.45} & \textbf{90.52} & \textbf{18.28} & 91.66  \\ 
 \cline{2-9} 
 &
  \multicolumn{1}{c|}{\multirow{4}{*}{DSelect\_k}} &
  \multicolumn{1}{c|}{MoCo} &
  99.99 &
  \multicolumn{1}{c|}{99.95} &
82.29                        & 89.78                        & 18.45                        & 91.22 \\ 
 &
  \multicolumn{1}{c|}{} &
  \multicolumn{1}{c|}{Aligned-MTL} &
  99.99 &
  \multicolumn{1}{c|}{99.95} &
80.60                        & 88.62                        & 18.85                        & 90.77 \\ 
 &
  \multicolumn{1}{c|}{} &
  \multicolumn{1}{c|}{STCH} &
  \textbf{99.99} &
  \multicolumn{1}{c|}{\textbf{99.95}} &
\textbf{83.46} & \textbf{90.52} & \textbf{18.22} & \textbf{91.73} \\ 
 &
  \multicolumn{1}{c|}{} &
  \multicolumn{1}{c|}{DB-MTL} &
  99.99 &
  \multicolumn{1}{c|}{99.95} &
83.37                        & 90.47                        & 18.26                        & 91.63  \\ 
 \cline{2-9} 
 &
  \multicolumn{1}{c|}{\multirow{4}{*}{TESLA}} &
  \multicolumn{1}{c|}{MoCo} &
  99.99 &
  \multicolumn{1}{c|}{99.95} &
84.09                        & 90.90                        & 17.94                        & 92.02\\ 
 &
  \multicolumn{1}{c|}{} &
  \multicolumn{1}{c|}{Aligned-MTL} &
  99.99 &
  \multicolumn{1}{c|}{99.95} &
84.10                        & 90.91                        & 18.00                        & 92.03   \\ 
 &
  \multicolumn{1}{c|}{} &
  \multicolumn{1}{c|}{STCH} &
  \textbf{99.99} &
  \multicolumn{1}{c|}{\textbf{99.95}} &
\textbf{85.09} & \textbf{91.53} & 17.76                        & \textbf{92.47}   \\ 
 &
  \multicolumn{1}{c|}{} &
  \multicolumn{1}{c|}{DB-MTL} &
  99.99 &
  \multicolumn{1}{c|}{99.95} &
84.88                        & 91.40                        & \textbf{17.72} & 92.39  \\ 
 \hline
\multirow{13}{*}{DINOv2}& \multicolumn{2}{c|}{Classification Only}  &99.99 & \multicolumn{1}{c|}{99.95} &  N/A  &N/A  &N/A  &N/A    \\

& \multicolumn{2}{c|}{Segmentation Only} & N/A &  \multicolumn{1}{c|}{N/A}  &88.19&  93.39& 15.22& 94.04 \\

\cline{2-9}
&  \multicolumn{1}{c|}{\multirow{4}{*}{MMoE}} &
  \multicolumn{1}{c|}{MoCo} &
  99.99 &
  \multicolumn{1}{c|}{99.95} &
90.33                        & 94.71                        & 14.25                        & 94.83\\ 
 &
  \multicolumn{1}{c|}{} &
  \multicolumn{1}{c|}{Aligned-MTL} &
  99.99 &
  \multicolumn{1}{c|}{99.95} &
89.69                        & 94.32                        & 14.51                        & 94.63 \\ 
 &
  \multicolumn{1}{c|}{} &
  \multicolumn{1}{c|}{STCH} &
  99.99 &
  \multicolumn{1}{c|}{99.95} &
91.47                        & 95.39                        & 13.75                        & 95.36 \\ 
 &
  \multicolumn{1}{c|}{} &
  \multicolumn{1}{c|}{DB-MTL} &
  \textbf{99.99} &
  \multicolumn{1}{c|}{\textbf{99.95}} &
\textbf{91.48} & \textbf{95.39} & \textbf{13.66} & \textbf{95.40} \\  \cline{2-9} 
 &
  \multicolumn{1}{c|}{\multirow{4}{*}{DSelect\_k}} &
  \multicolumn{1}{c|}{MoCo} &
  99.99 &
  \multicolumn{1}{c|}{99.95} &
90.08                        & 94.56                        & 14.44                        & 94.74 \\ 
 &
  \multicolumn{1}{c|}{} &
  \multicolumn{1}{c|}{Aligned-MTL} &
  99.99 &
  \multicolumn{1}{c|}{99.95} &
87.81                        & 93.14                        & 15.06                        & 93.95\\ 
 &
  \multicolumn{1}{c|}{} &
  \multicolumn{1}{c|}{STCH} &
  99.99 &
  \multicolumn{1}{c|}{99.95} &
90.81                        & 94.99                        & \textbf{13.92} & 95.13 \\ 
 &
  \multicolumn{1}{c|}{} &
  \multicolumn{1}{c|}{DB-MTL} &
  \textbf{99.99} &
  \multicolumn{1}{c|}{\textbf{99.95}} &
\textbf{91.08} & \textbf{95.15} & 13.93                        & \textbf{95.20} \\  \cline{2-9} 
 &
  \multicolumn{1}{c|}{\multirow{4}{*}{TESLA}} &
  \multicolumn{1}{c|}{MoCo} &
  99.99 &
  \multicolumn{1}{c|}{99.95} &
91.37                        & 95.33                        & 13.84                        & 95.25 \\ 
 &
  \multicolumn{1}{c|}{} &
  \multicolumn{1}{c|}{Aligned-MTL} &
  99.99 &
  \multicolumn{1}{c|}{99.95} &
90.11                        & 94.58                        & 14.37                        & 94.71  \\ 
 &
  \multicolumn{1}{c|}{} &
  \multicolumn{1}{c|}{STCH} &
  99.99 &
  \multicolumn{1}{c|}{99.95} &
91.65                        & 95.49                        & 13.48                        & 95.46  \\ 
 &
  \multicolumn{1}{c|}{} &
  \multicolumn{1}{c|}{DB-MTL} &
  \textbf{99.99} &
  \multicolumn{1}{c|}{\textbf{99.95}} &
\textbf{91.76} & \textbf{95.56} & \textbf{13.33} & \textbf{95.47}  \\  \hline

\multirow{3}{*}{EndoARSS} & 

\multicolumn{1}{c|}{\multirow{1}{*}{MMoE}} &  
  \multicolumn{1}{c|}{\multirow{3}{*}{DB-MTL}} &
  \multicolumn{1}{c}{99.99}&
  \multicolumn{1}{c|}{99.95} &
91.99 & 95.65 & 13.41 & 95.53 \\ \cline{2-2} 
&
\multicolumn{1}{c|}{\multirow{1}{*}{DSelect\_k}} &  

 &
   \multicolumn{1}{c}{99.99} &
  \multicolumn{1}{c|}{99.95} &
91.94 & 95.62 & 13.28 & 95.54  \\ \cline{2-2} 
&
\multicolumn{1}{c|}{\multirow{1}{*}{TESLA}} &  

 &
  \multicolumn{1}{c}{\textbf{99.99}} &
  \multicolumn{1}{c|}{\textbf{99.95}} &
\textbf{92.16} & \textbf{95.79} & \textbf{13.25} & \textbf{95.67}\\ \bottomrule[1pt] 

\end{tabular}}
\end{table*}

\begin{table*}
\renewcommand{\arraystretch}{1}
\caption{Multi-task learning results on the MTLEndovis dataset. We combine and compare the selected components one by one with our EndoARSS. Since STCH is the best-performing architecture in the MTLEndovis dataset, we only provide the results of the EndoARSS backbone (DINOv2 with SMA) based on MTLEndovis for comparison with other SOTA combinations.}
\label{tab:MTLEndovis}
\centering
\resizebox{\textwidth}{!}{
\begin{tabular}{c|cc|cc|cccc}
\toprule[1pt]
\multirow{2}{*}{Backbone} &
  \multicolumn{2}{c|}{Multi-Task Algorithms} &
  \multicolumn{2}{c|}{Activity Recognition} &
  \multicolumn{4}{c}{Semantic Segmentation} \\ \cline{2-9} 
 &
 \multicolumn{1}{c|}{Architectures} &
 \multicolumn{1}{c|}{\makecell[c]{Optimization \\ Strategies}} &
  AC(\%) &
  \multicolumn{1}{c|}{F-score(\%)} &
  mIoU(\%) &
  DS(\%) &
  HD &
  SSIM(\%) \\ \hline
\multirow{12}{*}{ResNet101} &
  \multicolumn{1}{c|}{\multirow{4}{*}{MMoE}} &
  MoCo &
  63.35 &
  54.62 &
  54.24 &
  68.40 &
  24.02 &
  86.47 \\ 
 &
  \multicolumn{1}{c|}{} &
  Aligned-MTL &
  62.61 &
  54.68 &
  57.68 &
  70.13 &
  \textbf{21.62} &
  \textbf{88.86} \\ 
 &
  \multicolumn{1}{c|}{} &
  STCH &
  65.80 &
  55.67 &
  56.30 &
  68.40 &
  21.94 &
  87.20\\ 
 &
  \multicolumn{1}{c|}{} &
  DB-MTL &
  \textbf{65.83} &
  \textbf{57.52} &
  \textbf{58.45} &
  \textbf{72.45} &
  22.04 &
  88.52 \\ \cline{2-9} 
 &
  \multicolumn{1}{c|}{\multirow{4}{*}{DSelect\_k}} &
  MoCo &
  66.83 &
  57.69 &
  54.53 &
  68.74 &
  24.10 &
  85.64 \\ 
 &
  \multicolumn{1}{c|}{} &
  Aligned-MTL &
  66.31 &
  56.58 &
  \textbf{61.24} &
  72.24 &
  \textbf{21.71} &
  \textbf{89.77} \\ 
 &
  \multicolumn{1}{c|}{} &
  STCH &
  65.82 &
  58.48 &
  58.13 &
  71.71 &
  23.01 &
  88.06 \\ 
 &
  \multicolumn{1}{c|}{} &
  DB-MTL &
  \textbf{66.89} &
  \textbf{59.12} &
  60.62 &
  \textbf{74.10} &
  22.36 &
  88.41 \\ \cline{2-9} 
 &
  \multicolumn{1}{c|}{\multirow{4}{*}{TESLA}} &
  MoCo &
  67.12 &
  61.23 &
  62.14 &
  74.17 &
  21.58 &
  87.44 \\ 
 &
  \multicolumn{1}{c|}{} &
  Aligned-MTL &
  \textbf{67.94} &
  61.46 &
  64.13 &
  75.14 &
  21.54 &
  \textbf{89.03} \\ 
 &
  \multicolumn{1}{c|}{} &
  STCH &
  67.23 &
  \textbf{61.76} &
  \textbf{64.26} &
  \textbf{77.16} &
  \textbf{21.52} &
  88.70 \\ 
 &
  \multicolumn{1}{c|}{} &
  DB-MTL &
  67.51 &
  61.42 &
  63.31 &
  76.21 &
  21.73 &
  88.69 \\ \hline
\multirow{12}{*}{DINOv2} & \multicolumn{2}{c|}{Classification Only} &68.90 & \multicolumn{1}{c|}{62.52} &  N/A  &N/A  &N/A  &N/A    \\
&
\multicolumn{2}{c|}{Segmentation Only} & N/A &  \multicolumn{1}{c|}{N/A}  &63.34&  76.53& 21.27& 88.93 \\ 

\cline{2-9}
&
  \multicolumn{1}{c|}{\multirow{4}{*}{MMoE}} &
  MoCo &
  70.56 &
  65.41 &
  64.93 &
  77.78 &
  21.11 &
  89.15 \\ 
 &
  \multicolumn{1}{c|}{} &
  Aligned-MTL &
  70.14 &
  64.81 &
  65.57 &
  77.98 &
  21.37 &
  89.13 \\ 
 &
  \multicolumn{1}{c|}{} &
  STCH &
  \textbf{71.66} &
  \textbf{65.71} &
  \textbf{66.25} &
  \textbf{78.65} &
  \textbf{20.67} &
  89.55 \\ 
 &
  \multicolumn{1}{c|}{} &
  DB-MTL &
  71.07 &
  65.61 &
  66.15 &
  78.42 &
  20.85 &
  \textbf{89.83} \\ \cline{2-9} 
 &
  \multicolumn{1}{c|}{\multirow{4}{*}{DSelect\_k}} &
  MoCo &
  69.95 &
  65.51 &
  66.90 &
  78.98 &
  20.99 &
  89.84 \\ 
 &
  \multicolumn{1}{c|}{} &
  Aligned-MTL &
  69.14 &
  \textbf{65.80} &
  67.67 &
  79.53 &
  20.47 &
  89.88 \\ 
 &
  \multicolumn{1}{c|}{} &
  STCH &
  \textbf{71.73} &
  65.61 &
  \textbf{68.19} &
  \textbf{79.88} &
  \textbf{20.35} &
  \textbf{90.07} \\ 
 &
  \multicolumn{1}{c|}{} &
  DB-MTL &
  69.75 &
  65.20 &
  67.97 &
  79.78 &
  20.69 &
  89.82 \\ \cline{2-9} 
 &
  \multicolumn{1}{c|}{\multirow{4}{*}{TESLA}} &
  MoCo &
  71.01 &
  66.29 &
  68.39 &
  80.99 &
  20.22 &
  90.22 \\ 
 &
  \multicolumn{1}{c|}{} &
  Aligned-MTL &
  71.80 &
  \textbf{66.57} &
  68.89 &
  80.58 &
  20.26 &
  89.83 \\ 
 &
  \multicolumn{1}{c|}{} &
  STCH &
  \textbf{71.92} &
  66.46 &
  \textbf{70.64} &
  \textbf{81.77} &
  \textbf{20.00} &
  \textbf{90.84} \\ 
 &
  \multicolumn{1}{c|}{} &
  DB-MTL &
  71.78 &
  65.98 &
  68.36 &
  78.57 &
  20.17 &
  90.11 \\ \hline
\multirow{3}{*}{EndoARSS} &
  \multicolumn{1}{c|}{MMoE} &
  \multirow{3}{*}{STCH} &
  71.98 &
 \textbf{ 66.86} &
  68.28 &
  79.98 &
  20.25 &
  90.12 \\ \cline{2-2}  
 &
  \multicolumn{1}{c|}{DSelect\_k} &
   &
  71.49 &
  66.29 &
  68.49 &
  80.61 &
  20.18 &
  90.85 \\ \cline{2-2}  
 &
  \multicolumn{1}{c|}{TESLA} &
   &
  \textbf{72.18} &
           66.77 &
  \textbf{72.95} &
  \textbf{83.47} &
  \textbf{19.52} &
  \textbf{91.26} \\ 
  \bottomrule[1pt]
\end{tabular}}
\end{table*}

\subsubsection{Evaluation on MTLEndovis}
MTLEndovis poses significant challenges since it contains a higher number of classification and segmentation categories and more complex surgical scenes than the MTLESD dataset. Table~\ref{tab:MTLEndovis} shows the quantitative comparison results. Despite these difficulties, our proposed EndoARSS framework exhibits remarkable performance improvements and exceptional generalization capabilities, demonstrating its strong surgical scenario-understanding capability. Compared to the ResNet101 and DINOv2 based methods, EndoARSS consistently outperforms across all key metrics, including both tasks. On the activity recognition task, the accuracy of EndoARSS backbone with TESLA reaches 72.18\%, a clear improvement of 0.36\%-8.83\% over other methods. This is especially significant given the increased complexity of classification in MTLEndovis, with a larger variety of surgical activities and instruments. In terms of segmentation, our framework achieves a notable mIoU of 72.95\%. DS also sees a significant maximum increase of 15.07\%, while HD is reduced by 4.5, indicating enhanced boundary precision. SSIM improves to 91.26\%, highlighting the effectiveness in maintaining fine-grained structural details.

\subsubsection{Evaluation on MTLEndovis-Gen}
To further assess the generalization capability of our model, we evaluate it on the MTLEndovis-Gen dataset after training on the MTLEndovis dataset. As shown in Table~\ref{tab:MTLEndovis-Gen}, EndoARSS achieves the best overall performance, with the TESLA strategy reaching 49.24\% in classification accuracy and 49.10\% mIoU in segmentation. Notably, it maintains strong boundary precision with the lowest HD (22.17) and highest SSIM (88.27), validating the model’s capacity to generalize to unseen surgical scenarios. This highlights the strength of our multi-task framework in capturing transferable representations that remain effective under domain shift, which is crucial for practical deployment in real-world surgical environments.

\begin{table*}[]
\renewcommand{\arraystretch}{1}
\caption{Multi-task learning results on the MTLEndovis-Gen dataset. The model setup is the same as the MTLEndovis dataset evaluation.}
\label{tab:MTLEndovis-Gen}
\centering
\resizebox{\textwidth}{!}{
\begin{tabular}{c|cc|cccccc}
\toprule[1pt]
\multirow{4}{*}{Backbone}   & \multicolumn{2}{c|}{\multirow{2}{*}{Multi-Task Algorithms}}                                                                              & \multicolumn{2}{c|}{\multirow{2}{*}{Classification}}                      & \multicolumn{4}{c}{\multirow{2}{*}{Segmentation}}                                           \\
                            & \multicolumn{2}{c|}{}                                                                                                                    & \multicolumn{2}{c|}{}                                                     & \multicolumn{4}{c}{}                                                                        \\ \cline{2-9} 
                            & \multicolumn{1}{c|}{\multirow{2}{*}{\begin{tabular}[c]{@{}c@{}}Optimization\\ Strategies\end{tabular}}} & \multirow{2}{*}{Architectures} & \multirow{2}{*}{Accuracy} & \multicolumn{1}{c|}{\multirow{2}{*}{F-score}} & \multirow{2}{*}{mIoU} & \multirow{2}{*}{Dice} & \multirow{2}{*}{HD} & \multirow{2}{*}{SSIM} \\
                            & \multicolumn{1}{c|}{}                                                                                   &                                &                           & \multicolumn{1}{c|}{}                         &                       &                       &                     &                       \\ \hline
\multirow{12}{*}{ResNet101} & \multicolumn{1}{c|}{\multirow{4}{*}{MMoE}}                                                              & MoCo                           & \textbf{42.59}          & \multicolumn{1}{c|}{\textbf{42.77}}         & 42.40               & 41.48               & 27.39               & 84.76               \\
                            & \multicolumn{1}{c|}{}                                                                                   & Aligned-MTL                    & 42.07                   & \multicolumn{1}{c|}{40.81}                  & \textbf{43.85}      & 40.14               & 27.55               & 83.51               \\
                            & \multicolumn{1}{c|}{}                                                                                   & \multicolumn{1}{c|}{STCH}      & 42.37                   & \multicolumn{1}{c|}{40.49}                  & 43.82               & 41.50               & \textbf{26.59}      & 84.21               \\
                            & \multicolumn{1}{c|}{}                                                                                   & DB-MTL                         & 41.40                   & \multicolumn{1}{c|}{42.60}                  & 43.20               & \textbf{42.17}      & 26.76               & \textbf{85.84}      \\ \cline{2-9} 
                            & \multicolumn{1}{c|}{\multirow{4}{*}{DSelect\_k}}                                                        & MoCo                           & 41.52                   & \multicolumn{1}{c|}{43.84}                  & 42.79               & 41.14               & 27.11               & 84.43               \\
                            & \multicolumn{1}{c|}{}                                                                                   & Aligned-MTL                    & 42.57                   & \multicolumn{1}{c|}{43.93}                  & 43.06               & 40.14               & 27.82               & 84.13               \\
                            & \multicolumn{1}{c|}{}                                                                                   & \multicolumn{1}{c|}{STCH}      & \textbf{43.00}          & \multicolumn{1}{c|}{\textbf{44.42}}         & 42.88               & 40.03               & \textbf{26.31}      & \textbf{85.00}      \\
                            & \multicolumn{1}{c|}{}                                                                                   & DB-MTL                         & 42.34                   & \multicolumn{1}{c|}{43.99}                  & \textbf{45.54}      & \textbf{41.62}      & 27.07               & 84.26               \\ \cline{2-9} 
                            & \multicolumn{1}{c|}{\multirow{4}{*}{TESLA}}                                                             & MoCo                           & 44.94                   & \multicolumn{1}{c|}{44.96}                  & 44.64               & 42.61               & 26.8                & 84.00               \\
                            & \multicolumn{1}{c|}{}                                                                                   & Aligned-MTL                    & 44.09                   & \multicolumn{1}{c|}{44.83}                  & 46.43               & 43.19               & 25.92               & 85.99               \\
                            & \multicolumn{1}{c|}{}                                                                                   & \multicolumn{1}{c|}{STCH}      & 44.32                   & \multicolumn{1}{c|}{45.97}                  & \textbf{46.54}      & 42.64               & \textbf{25.03}      & \textbf{86.34}      \\
                            & \multicolumn{1}{c|}{}                                                                                   & DB-MTL                         & \textbf{45.27}          & \multicolumn{1}{c|}{\textbf{46.46}}                              & 46.44               & \textbf{43.53}      & 26.61               & 85.82               \\ \hline
\multirow{12}{*}{Dinov2}    & \multicolumn{1}{c|}{\multirow{4}{*}{MMoE}}                                                              & MoCo                           & 45.64                   & \multicolumn{1}{c|}{47.22}                  & 46.63               & 45.65               & 24.64               & \textbf{86.97}      \\
                            & \multicolumn{1}{c|}{}                                                                                   & Aligned-MTL                    & 45.59                   & \multicolumn{1}{c|}{46.52}                  & \textbf{47.75}      & 46.69               & \textbf{24.82}      & 85.69               \\
                            & \multicolumn{1}{c|}{}                                                                                   & \multicolumn{1}{c|}{STCH}      & \textbf{46.21}          & \multicolumn{1}{c|}{46.29}                  & 47.11               & \textbf{47.46}      & 24.79               & 85.36               \\
                            & \multicolumn{1}{c|}{}                                                                                   & DB-MTL                         & 45.65                   & \multicolumn{1}{c|}{\textbf{47.65}}         & 46.72               & 46.38               & 24.77               & 85.62               \\ \cline{2-9} 
                            & \multicolumn{1}{c|}{\multirow{4}{*}{DSelect\_k}}                                                        & MoCo                           & 46.98                   & \multicolumn{1}{c|}{46.47}                  & 47.19               & 46.17               & \textbf{24.98}      & 86.09               \\
                            & \multicolumn{1}{c|}{}                                                                                   & Aligned-MTL                    & 46.39                   & \multicolumn{1}{c|}{47.64}                  & 46.53               & 46.67               & 23.69               & 85.56               \\
                            & \multicolumn{1}{c|}{}                                                                                   & \multicolumn{1}{c|}{STCH}      & \textbf{47.01}          & \multicolumn{1}{c|}{46.07}                  & 46.52               & 47.20               & 24.4                & 85.73               \\
                            & \multicolumn{1}{c|}{}                                                                                   & DB-MTL                         & 46.39                   & \multicolumn{1}{c|}{\textbf{48.64}}         & \textbf{47.53}      & \textbf{47.67}      & 23.69               & \textbf{86.56}      \\ \cline{2-9} 
                            & \multicolumn{1}{c|}{\multirow{4}{*}{TESLA}}                                                             & MoCo                           & 48.28                   & \multicolumn{1}{c|}{50.99}                  & 47.20               & 48.74               & 22.51               & 87.11               \\
                            & \multicolumn{1}{c|}{}                                                                                   & Aligned-MTL                    & 48.96                   & \multicolumn{1}{c|}{50.11}                  & 48.31               & 49.50               & 22.11               & 87.37               \\
                            & \multicolumn{1}{c|}{}                                                                                   & \multicolumn{1}{c|}{STCH}      & \textbf{49.05}          & \multicolumn{1}{c|}{\textbf{51.24}}         & \textbf{48.82}      & \textbf{49.93}      & 22.13               & \textbf{88.25}      \\
                            & \multicolumn{1}{c|}{}                                                                                   & DB-MTL                         & 48.73                   & \multicolumn{1}{c|}{51.06}                  & 48.42               & 49.61               & \textbf{23.85}      & 87.55               \\ \hline
\multirow{3}{*}{EndoARSS}   & \multicolumn{1}{c|}{MMoE}                                                                               & \multirow{3}{*}{STCH}          & 47.19                   & \multicolumn{1}{c|}{48.14}                  & 47.71               & 48.29               & 22.4                & 86.90               \\ \cline{2-2}
                            & \multicolumn{1}{c|}{DSelect\_k}                                                                         &                                & 47.33                   & \multicolumn{1}{c|}{49.93}                  & 47.67               & 48.63               & 22.85               & 86.94               \\ \cline{2-2}
                            & \multicolumn{1}{c|}{TESLA}                                                                              &                                & \textbf{49.24}          & \multicolumn{1}{c|}{\textbf{52.78}}         & \textbf{49.10}      & \textbf{49.73}      & \textbf{22.17}      & \textbf{88.27}      \\ \bottomrule[1pt]
\end{tabular}}
\end{table*}


\subsubsection{Ablation Studies}
In order to explore the effect of proposed components in the EndoARSS framework, we conduct the ablation study on MTLESD and MTLEndovis datasets, and the results are shown in Table~\ref{tab:abl}. We decompose the framework into DINOv2 (backbone), TESLA (architecture), and SMA, and evaluate their individual contributions. Since the SMA can only work by attaching to the backbone of the DINOv2, there are no results for SMA without DINOv2. The results indicate that without DINOv2, the model struggles to capture high-quality visual features, leading to a significant drop in segmentation performance. This confirms that DINOv2's powerful feature extraction is key to handling the complexity of surgical scenes. The removal of TESLA results in reduced effectiveness in handling multi-task learning, as the model fails to isolate task-specific parameters efficiently. Similarly, without SMA, the framework struggles with discriminating between visually similar features in challenging surgical environments. 
Moreover, the relatively small standard deviations across different configurations suggest that our model's behavior is statistically reliable. This consistency strengthens the validity of the performance gains and supports the robustness of the proposed framework, especially in the context of multi-task learning in complex surgical scenes.
Overall, the integration of all three components yields the best performance, underscoring their complementary roles in achieving SOTA performance. 

To investigate the effect of the rank size on the MTL performance of our TESLA module, we conduct a series of comparative experiments with four different rank sizes. As shown in Table~\ref{tab:abl1}, the performance of EndoARSS improves as the rank increases from 1 to 4, but begins to degrade with further increases. This indicates that while a moderate rank provides sufficient capacity for TESLA to learn task-specific adaptations, excessively large ranks may introduce redundant or noisy parameter updates that can disturb the shared backbone features, ultimately degrading overall performance.

\begin{figure}[!t]
\centering
\includegraphics[width=0.48\textwidth]{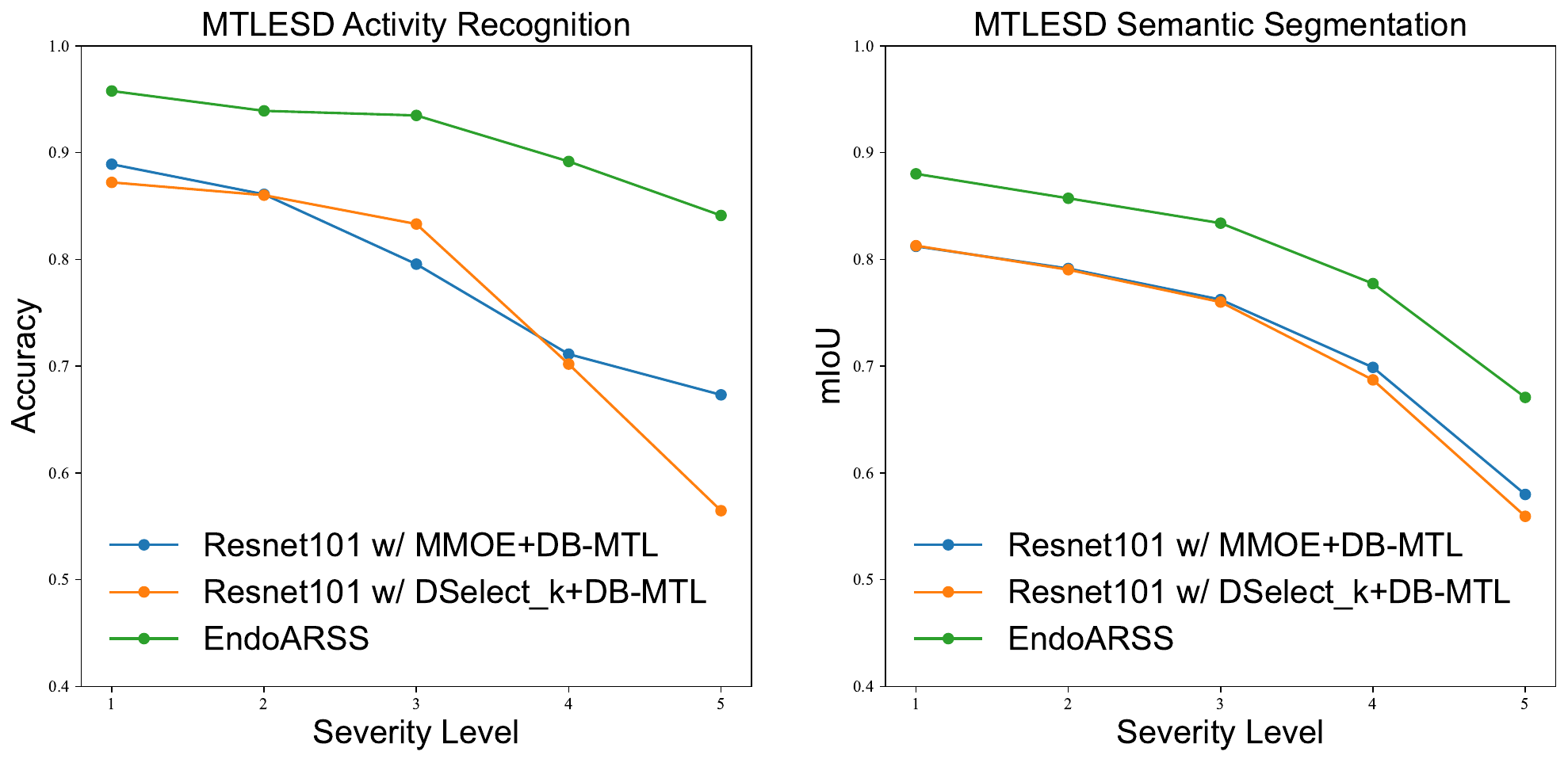}
\caption{Comparison of robustness across severity levels. We evaluate the robustness of different MTL models by applying nineteen distinct image corruption methods, each characterized by five levels of severity. We average the results of 5 levels, enabling a comprehensive assessment.}
\label{fig_robust}
\end{figure}

\begin{table*}
\caption{Ablation study on MTLESD and MTLEndovis datasets. The initial MTL model configuration in the first row is ResNet101 as the backbone, MMoE as the architecture, and DB-MTL as the optimization strategy. We add our proposed methods one by one to explore their effectiveness on the segmentation performance.}
\label{tab:abl}
\centering
\resizebox{\textwidth}{!}{
\renewcommand{\arraystretch}{1}
\begin{tabular}{c|ccc|cc|cccc}
\toprule[1pt]

\multirow{2}{*}{Dataset}    & \multirow{2}{*}{DINOv2} & \multirow{2}{*}{TESLA} & \multirow{2}{*}{SMA} & \multicolumn{2}{c|}{Activity Recognition} & \multicolumn{4}{c}{Semantic Segmentation} \\ \cline{5-10} 
                            &                         &                        &                      & Accuracy(\%)             & F-score(\%)            & mIoU(\%)      & Dice(\%)      & HD     & SSIM(\%)     \\ \hline
\multirow{6}{*}{MTLESD}     & $\times$                       & $\times$                      & $\times$                    & 99.99±0.01              & 99.95            & 83.45±6.41   & 90.52   & 18.28  & 91.66  \\
                            & $\checkmark$                       & $\times$                      & $\times$                    & 99.99±0.01              & 99.95            & 91.48±5.22   & 95.39   & 13.66  & 95.40  \\
                            & $\times$                       & $\checkmark$                      & $\times$                    & 99.99±0.01              & 99.95            & 84.88±6.32   & 91.40   & 17.72  & 92.39  \\
                            & $\checkmark$                       & $\checkmark$                      & $\times$                    & 99.99±0.01              & 99.95            & 91.76±5.18   & 95.56   & 13.33  & 95.47  \\
                            & $\checkmark$                       & $\times$                      & $\checkmark$                    & 99.99±0.01              & 99.95            & 91.99±5.06   & 95.65   & 13.41  & 95.53  \\
                            & $\checkmark$                       & $\checkmark$                      & $\checkmark$                    & \textbf{99.99±0.01}              & \textbf{99.95}            & \textbf{92.16±5.06}   & \textbf{95.79}   & \textbf{13.25}  & \textbf{95.67}  \\ \hline
\multirow{6}{*}{MTLEndovis} & $\times$                       & $\times$                      & $\times$                    & 65.83±29.80              & 57.52            & 58.45±12.75   & 72.45   & 22.04  & 88.52  \\
                            & $\checkmark$                       & $\times$                      & $\times$                    & 71.07±20.01              & 65.61            & 66.15±11.06   & 78.42   & 20.85  & 89.83  \\
                            & $\times$                       & $\checkmark$                      & $\times$                    & 67.51±25.05              & 61.42            & 63.31±9.80   & 76.21   & 21.73  & 88.69  \\
                            & $\checkmark$                       & $\checkmark$                      & $\times$                    & 71.78±18.33              & 65.98            & 68.36±9.11   & 78.57   & 20.17  & 90.11  \\
                            & $\checkmark$                       & $\times$                      & $\checkmark$                    & 71.98±21.90              & \textbf{66.86}            & 68.28±4.46   & 79.98   & 20.25  & 90.12  \\
                            & $\checkmark$                       & $\checkmark$                      & $\checkmark$                    & \textbf{72.18±21.07}              & 66.77            & \textbf{72.95±4.42}   & \textbf{83.47}   & \textbf{19.53}  & \textbf{91.26}  \\ \bottomrule[1pt]
\end{tabular}}
\end{table*}

\begin{table*}[]
\caption{Ablation study on the rank size of the task efficient shared low-rank adapters.}
\label{tab:abl1}
\resizebox{\textwidth}{!}{
\renewcommand{\arraystretch}{0.9}
\begin{tabular}{c|c|cc|cccc}
\toprule[1pt]
\multirow{2}{*}{Dataset}    & \multirow{2}{*}{Rank size} & \multicolumn{2}{c}{Activity Recognition} & \multicolumn{4}{c}{Semantic Segmentation}                         \\ \cline{3-8} 
                            &                            & Accuracy(\%)        & F-score(\%)        & mIoU(\%)       & Dice(\%)       & HD             & SSIM(\%)       \\ \hline
\multirow{4}{*}{MTLESD}     & 1                          & 99.99               & 99.95              & 91.89          & 95.62          & 13.41          & 95.49          \\
                            & 4                          & \textbf{99.99}      & \textbf{99.95}     & \textbf{92.16} & \textbf{95.79} & 13.25 & \textbf{95.67} \\
                            & 8                          & 99.99               & 99.95              & 91.95          & 95.65          & \textbf{13.20}          & 95.64          \\
                            & 16                         & 99.99               & 99.95              & 91.70          & 95.52          & 13.63          & 95.49          \\ \hline
\multirow{4}{*}{MTLEndovis} & 1                          & 71.89               & 66.05              & 68.84          & 80.40          & 20.03          & 90.42          \\
                            & 4                          & \textbf{72.18}      & \textbf{66.77}     & \textbf{72.95} & \textbf{83.47} & \textbf{19.53} & \textbf{91.26} \\
                            & 8                          & 72.01               & 66.23              & 69.51          & 80.98          & 20.02          & 90.46          \\
                            & 16                         & 72.04               & 66.12              & 69.15          & 80.74          & 20.42          & 90.43          \\ \bottomrule[1pt]
\end{tabular}}
\end{table*}

\begin{table*}[]
\caption{Robustness comparison on Resnet101 with DSelect\_k, Resnet101 with MMoE, and our EndoARSS. All MTL models utilize DB-MTL as the optimization strategy. We compute the average results on each corruption type.}
\label{tab:robust}
\centering
\resizebox{0.95\textwidth}{!}{
\renewcommand{\arraystretch}{1}
\begin{tabular}{cc|ccc|ccc}
\toprule[1pt]
\multicolumn{2}{c|}{Metric}                                   & \multicolumn{3}{c|}{AC(\%)}                                                                                           & \multicolumn{3}{c}{mIoU(\%)}                                                                                                      \\ \hline
\multicolumn{2}{c|}{model}                                    & \multicolumn{1}{c|}{Res w/ DS} & \multicolumn{1}{c|}{Res w/ MMOE} & Ours    & \multicolumn{1}{c|}{Res w/ DS} & \multicolumn{1}{c|}{Res w/ MMOE} & Ours                  \\ \hline
\multicolumn{1}{c|}{}                            & Gauss.     & \multicolumn{1}{c|}{79.89}                        & \multicolumn{1}{c|}{98.95}               & \textbf{100.00} & \multicolumn{1}{c|}{59.61}                     & \multicolumn{1}{c|}{60.85}               & \textbf{66.87}                        \\  
\multicolumn{1}{c|}{}                            & Shot       & \multicolumn{1}{c|}{78.58}                        & \multicolumn{1}{c|}{96.95}               & \textbf{100.00} & \multicolumn{1}{c|}{64.18}                     & \multicolumn{1}{c|}{65.90}               & \textbf{83.22}                        \\  
\multicolumn{1}{c|}{}                            & Impulse    & \multicolumn{1}{c|}{74.13}                        & \multicolumn{1}{c|}{\textbf{99.80}}      & 99.70           & \multicolumn{1}{c|}{57.02}                     & \multicolumn{1}{c|}{58.38}               & \textbf{78.59}                        \\  
\multicolumn{1}{c|}{\multirow{-4}{*}{Noise}}     & Speckle    & \multicolumn{1}{c|}{99.78}                        & \multicolumn{1}{c|}{100.00}              & \textbf{100.00} & \multicolumn{1}{c|}{72.50}                     & \multicolumn{1}{c|}{74.53}               & \textbf{86.69}                        \\ \hline
\multicolumn{1}{c|}{}                            & Defocus    & \multicolumn{1}{c|}{76.04}                        & \multicolumn{1}{c|}{72.21}               & \textbf{85.79}  & \multicolumn{1}{c|}{74.50}                     & \multicolumn{1}{c|}{75.45}               & \textbf{80.78}                        \\  
\multicolumn{1}{c|}{}                            & Gauss.     & \multicolumn{1}{c|}{73.16}                        & \multicolumn{1}{c|}{65.61}               & \textbf{86.21}  & \multicolumn{1}{c|}{73.23}                     & \multicolumn{1}{c|}{74.03}               & \textbf{79.06}                        \\  
\multicolumn{1}{c|}{}                            & Glass      & \multicolumn{1}{c|}{80.58}                        & \multicolumn{1}{c|}{73.04}               & \textbf{99.12}  & \multicolumn{1}{c|}{77.25}                     & \multicolumn{1}{c|}{77.78}               & \textbf{82.11}                        \\  
\multicolumn{1}{c|}{}                            & Motion     & \multicolumn{1}{c|}{73.59}                        & \multicolumn{1}{c|}{71.72}               & \textbf{91.45}  & \multicolumn{1}{c|}{71.34}                     & \multicolumn{1}{c|}{73.43}               & \textbf{76.73}                        \\  
\multicolumn{1}{c|}{\multirow{-5}{*}{Blur}}      & Zoom       & \multicolumn{1}{c|}{78.13}                        & \multicolumn{1}{c|}{61.95}               & \textbf{98.19}  & \multicolumn{1}{c|}{72.19}                     & \multicolumn{1}{c|}{72.54}               & \textbf{74.99}                        \\ \hline
\multicolumn{1}{c|}{}                            & Bleeding   & \multicolumn{1}{c|}{96.12}                        & \multicolumn{1}{c|}{96.16}               & \textbf{97.22}  & \multicolumn{1}{c|}{75.84}                     & \multicolumn{1}{c|}{75.12}               &  \textbf{77.02} \\  
\multicolumn{1}{c|}{}                            & Brightness & \multicolumn{1}{c|}{83.68}                        & \multicolumn{1}{c|}{84.04}               & \textbf{97.53}  & \multicolumn{1}{c|}{81.14}                     & \multicolumn{1}{c|}{81.29}               & \textbf{85.06}                        \\  
\multicolumn{1}{c|}{}                            & Smoke        & \multicolumn{1}{c|}{8.71}                         & \multicolumn{1}{c|}{8.71}                & \textbf{50.53}  & \multicolumn{1}{c|}{62.49}                     & \multicolumn{1}{c|}{66.24}               & \textbf{69.03}                        \\  
\multicolumn{1}{c|}{\multirow{-4}{*}{Occlusion}} & Spatter    & \multicolumn{1}{c|}{98.06}                        & \multicolumn{1}{c|}{97.06}               & \textbf{100.00} & \multicolumn{1}{c|}{77.75}                     & \multicolumn{1}{c|}{77.38}               & \textbf{81.48}                        \\ \hline
\multicolumn{1}{c|}{}                            & Contrast   & \multicolumn{1}{c|}{8.71}                         & \multicolumn{1}{c|}{8.71}                & \textbf{29.85}  & \multicolumn{1}{c|}{60.43}                     & \multicolumn{1}{c|}{63.75}               & \textbf{66.41}                        \\  
\multicolumn{1}{c|}{}                            & Elastic    & \multicolumn{1}{c|}{98.35}                        & \multicolumn{1}{c|}{\textbf{99.44}}      & 99.10           & \multicolumn{1}{c|}{65.22}                     & \multicolumn{1}{c|}{62.97}               & \textbf{68.58}                        \\  
\multicolumn{1}{c|}{}                            & Gamma      & \multicolumn{1}{c|}{100.00}                       & \multicolumn{1}{c|}{99.27}               & \textbf{100.00} & \multicolumn{1}{c|}{82.76}                     & \multicolumn{1}{c|}{82.89}               & \textbf{89.43}                        \\  
\multicolumn{1}{c|}{}                            & Jpeg       & \multicolumn{1}{c|}{95.59}                        & \multicolumn{1}{c|}{92.74}               & \textbf{100.00} & \multicolumn{1}{c|}{79.55}                     & \multicolumn{1}{c|}{79.40}               & \textbf{90.16}                        \\  
\multicolumn{1}{c|}{}                            & Pixelate   & \multicolumn{1}{c|}{100.00}                       & \multicolumn{1}{c|}{100.00}              & \textbf{100.00} & \multicolumn{1}{c|}{81.94}                     & \multicolumn{1}{c|}{82.13}               & \textbf{89.56}                        \\  
\multicolumn{1}{c|}{\multirow{-6}{*}{Digital}}   & Saturate   & \multicolumn{1}{c|}{52.95}                        & \multicolumn{1}{c|}{66.89}               & \textbf{99.78}  & \multicolumn{1}{c|}{80.29}                     & \multicolumn{1}{c|}{80.73}               & \textbf{86.05}                        \\ \hline
\multicolumn{2}{c|}{Mean}                                     & \multicolumn{1}{c|}{76.63}                        & \multicolumn{1}{c|}{78.59}               & \textbf{91.29}  & \multicolumn{1}{c|}{72.06}                     & \multicolumn{1}{c|}{72.88}               & \textbf{79.57}                        \\ \bottomrule[1pt]
\end{tabular}}
\end{table*}

\subsubsection{Robustness Experiments}

We further assess the robustness of our EndoARSS framework on the MTLESD dataset by evaluating its performance under various image corruptions. Following the configuration in~\cite{bai2025surgical}, we introduce 19 types of image corruptions at 5 severity levels, encompassing noise (Gaussian noise, shot noise, impulse noise, speckle noise), blurs (defocus blur, glass blur, motion blur, zoom blur), occlusion (bleeding, brightness adjustment, smoke, spatter), and digital (contrast adjustment, elastic transformation, gamma correction, JPEG compression, pixelation, saturation). These simulated degradation scenarios cover multiple factors affecting vision and reproduce the complex visual conditions in real surgeries. We conduct inference on the corrupted test dataset and compute the average results across different corruption types at each severity level. In Figure~\ref{fig_robust}, we can observe that the multi-task performance of MTL models degrades as corruption severity increases. EndoARSS consistently outperforms baseline models across all severity levels demonstrating superior robustness. Table~\ref{tab:robust} presents the detailed average results for each corruption type, where EndoARSS exhibits a marked advantage in handling diverse corruption scenarios.

\subsection{Discussion and limitations}

Overall, our EndoARSS framework demonstrates significant performance improvements across three datasets, indicating its ability to adapt to varied endoscopic scenarios. In particular, the introduction of the Spatially-Aware Multi-Scale Attention (SMA) module significantly enhances segmentation performance, reaching a mean Intersection over Union (mIoU) of 92.16\% on the MTLESD dataset, surpassing SOTA methods. The combination of TESLA with DINOv2 further reduces gradient conflicts, facilitating efficient parameter sharing and contributing to superior results in challenging endoscopic surgical environments. Notably, our approach maintains robust performance even under image corruptions across multiple severity levels, further supporting its potential for generalization. The model's ability to handle heterogeneous multi-task data, while maintaining high performance in both tasks, confirms the effectiveness of leveraging cross-task feature interactions in surgical tasks. However, certain challenging scenarios in the MTLEndovis dataset, which features more complex surgical scenes and higher number of classification categories, still present occasional drops in accuracy, especially in tasks involving intricate tool-tissue interactions.

Beyond performance gains, EndoARSS also exhibits notable computational efficiency. Although it incorporates components such as SMA and DINOv2, our implementation remains resource-friendly. Training on a single NVIDIA RTX 4090 GPU requires only 9858MiB of memory, and inference occupies just 1668MiB while achieving a speed of 35.66 images per second. These characteristics indicate that EndoARSS is well-suited for real-time or near-real-time surgical applications, even in moderately resourced environments. 

Nonetheless, some limitations remain. The reliance on large-scale datasets for fine-tuning DINOv2 may limit the applicability of our approach to real-time clinical environments. Additionally, while TESLA effectively mitigates gradient conflicts, further improvements could be made to enhance model performance in highly complex, multi-task surgical scenarios. Future work could focus on the construction of large-scale multi-task datasets that cover a broader range of endoscopic surgical scenarios. By incorporating more diverse anatomical structures, surgical tools, and procedure types, such datasets will enable more comprehensive evaluation of generalization and promote the development of clinically applicable models.


\section{Ethical and Societal Considerations}
The development of EndoARSS brings several important ethical and societal considerations. Foremost among these is patient safety. Our system is designed purely as an assistive tool to support surgeons in enhancing procedural accuracy; all final decisions remain firmly under the control of the human operator. The framework does not support, nor is it intended for, autonomous decision-making. For data privacy, all endoscopic videos used in this study were sourced from publicly available datasets and thoroughly de-identified in accordance with established medical data protection guidelines.

We also recognize the broader implications for clinical integration. The surgical environment understanding functionalities of EndoARSS are intended to ease cognitive demands by automating tasks such as semantic segmentation and surgical activity recognition. By efficiently adapting foundation models, the system has the potential to make advanced surgical AI accessible in settings with limited computational infrastructure. At the same time, we are mindful of the risk that such technologies, if not carefully disseminated, could deepen existing inequalities. To mitigate this, it is crucial to explore measures such as open-weight model releases and usability studies tailored to diverse clinical environments.

\section{Conclusion}
This paper presents EndoARSS, a novel multi-task learning framework tailored for surgical activity recognition and semantic segmentation tasks in endoscopic surgery. Built on the DINOv2 foundation model, EndoARSS utilizes Low-Rank Adaptation (LoRA) to efficiently fine-tune the model while introducing Task Efficient Shared Low-rank Adapters (TESLA) which mitigates gradient conflicts across tasks. Furthermore, the Spatially-Aware Multi-Scale Attention (SMA) module improves feature representation discrimination by cross-spatial learning of global information in complex surgical environments. Extensive experiments on our proposed three datasets show that our framework outperforms existing models, achieving superior performance in both tasks. These results demonstrate the effectiveness and generalization capability of EndoARSS in addressing the challenges of endoscopic surgery, providing valuable insights for the comprehensive understanding of surgical scenarios. The promising outcomes of this study suggest that EndoARSS could play a pivotal role in advancing AI-driven endoscopic surgical systems, ultimately improving surgical safety and efficiency.

In the future, we aim to further address the high computational costs, exploring lightweight architectures and adaptive training mechanisms to reduce memory consumption and enhance real-time applicability in complex, multi-task surgical scenarios. Besides, exploring the versatility of our framework for other surgical modalities and tasks will be conducted to extend its clinical applications.

\section{Acknowledgements}
This work is the results of the research projects by Hong Kong RGC RIF R4020-22, CRF C4026-21GF, GRF 14203323, GRF 14216022, and GRF 14211420.

\bibliographystyle{IEEEtran}
\bibliography{mybib}

\begin{thebibliography}{10}
\providecommand{\url}[1]{#1}
\csname url@samestyle\endcsname
\providecommand{\newblock}{\relax}
\providecommand{\bibinfo}[2]{#2}
\providecommand{\BIBentrySTDinterwordspacing}{\spaceskip=0pt\relax}
\providecommand{\BIBentryALTinterwordstretchfactor}{4}
\providecommand{\BIBentryALTinterwordspacing}{\spaceskip=\fontdimen2\font plus
\BIBentryALTinterwordstretchfactor\fontdimen3\font minus \fontdimen4\font\relax}
\providecommand{\BIBforeignlanguage}[2]{{%
\expandafter\ifx\csname l@#1\endcsname\relax
\typeout{** WARNING: IEEEtran.bst: No hyphenation pattern has been}%
\typeout{** loaded for the language `#1'. Using the pattern for}%
\typeout{** the default language instead.}%
\else
\language=\csname l@#1\endcsname
\fi
#2}}
\providecommand{\BIBdecl}{\relax}
\BIBdecl

\bibitem{wang2023rethinking}
G.~Wang, L.~Bai, Y.~Wu, T.~Chen, and H.~Ren, ``Rethinking exemplars for continual semantic segmentation in endoscopy scenes: Entropy-based mini-batch pseudo-replay,'' \emph{Computers in Biology and Medicine}, vol. 165, p. 107412, 2023.

\bibitem{ali2020endoscopy}
S.~Ali, N.~Ghatwary, B.~Braden, D.~Lamarque, A.~Bailey, S.~Realdon, R.~Cannizzaro, J.~Rittscher, C.~Daul, and J.~East, ``Endoscopy disease detection challenge 2020,'' \emph{arXiv preprint arXiv:2003.03376}, 2020.

\bibitem{odagiri2017complications}
H.~Odagiri and H.~Yasunaga, ``Complications following endoscopic submucosal dissection for gastric, esophageal, and colorectal cancer: a review of studies based on nationwide large-scale databases,'' \emph{Annals of Translational Medicine}, vol.~5, no.~8, 2017.

\bibitem{yamamoto2009endoscopic}
S.~Yamamoto, N.~Uedo, R.~Ishihara, N.~Kajimoto, H.~Ogiyama, Y.~Fukushima, S.~Yamamoto, Y.~Takeuchi, K.~Higashino, H.~Iishi \emph{et~al.}, ``Endoscopic submucosal dissection for early gastric cancer performed by supervised residents: assessment of feasibility and learning curve,'' \emph{Endoscopy}, pp. 923--928, 2009.

\bibitem{gao2022savanet}
H.~Gao, W.~Fan, L.~Qiu, X.~Yang, Z.~Li, X.~Zuo, Y.~Li, M.~Q.-H. Meng, and H.~Ren, ``Savanet: Surgical action-driven visual attention network for autonomous endoscope control,'' \emph{IEEE Transactions on Automation Science and Engineering}, vol.~20, no.~4, pp. 2655--2667, 2022.

\bibitem{allan2020endovis18}
M.~Allan, S.~Kondo, S.~Bodenstedt, S.~Leger, R.~Kadkhodamohammadi, I.~Luengo, F.~Fuentes, E.~Flouty, A.~Mohammed, M.~Pedersen \emph{et~al.}, ``2018 robotic scene segmentation challenge,'' \emph{arXiv preprint arXiv:2001.11190}, 2020.

\bibitem{islam2020learning}
M.~Islam, L.~Seenivasan, L.~C. Ming, and H.~Ren, ``Learning and reasoning with the graph structure representation in robotic surgery,'' in \emph{International Conference on Medical Image Computing and Computer-Assisted Intervention}.\hskip 1em plus 0.5em minus 0.4em\relax Springer, 2020, pp. 627--636.

\bibitem{wang2023domain}
G.~Wang, T.-A. Ren, J.~Lai, L.~Bai, and H.~Ren, ``Domain adaptive sim-to-real segmentation of oropharyngeal organs,'' \emph{Medical \& Biological Engineering \& Computing}, vol.~61, no.~10, pp. 2745--2755, 2023.

\bibitem{bai2023surgical}
L.~Bai, M.~Islam, L.~Seenivasan, and H.~Ren, ``Surgical-vqla: Transformer with gated vision-language embedding for visual question localized-answering in robotic surgery,'' \emph{arXiv preprint arXiv:2305.11692}, 2023.

\bibitem{seenivasan2022global}
L.~Seenivasan, S.~Mitheran, M.~Islam, and H.~Ren, ``Global-reasoned multi-task learning model for surgical scene understanding,'' \emph{IEEE Robotics and Automation Letters}, 2022.

\bibitem{psychogyios2023sar}
D.~Psychogyios, E.~Colleoni, B.~Van~Amsterdam, C.-Y. Li, S.-Y. Huang, Y.~Li, F.~Jia, B.~Zou, G.~Wang, Y.~Liu \emph{et~al.}, ``Sar-rarp50: Segmentation of surgical instrumentation and action recognition on robot-assisted radical prostatectomy challenge,'' \emph{arXiv preprint arXiv:2401.00496}, 2023.

\bibitem{wang2024copesd}
G.~Wang, H.~Xiao, H.~Gao, R.~Zhang, L.~Bai, X.~Yang, Z.~Li, H.~Li, and H.~Ren, ``Copesd: A multi-level surgical motion dataset for training large vision-language models to co-pilot endoscopic submucosal dissection,'' \emph{arXiv preprint arXiv:2410.07540}, 2024.

\bibitem{tan2024endoood}
Q.~Tan, L.~Bai, G.~Wang, M.~Islam, and H.~Ren, ``Endoood: Uncertainty-aware out-of-distribution detection in capsule endoscopy diagnosis,'' \emph{arXiv preprint arXiv:2402.11476}, 2024.

\bibitem{bai2024ossar}
L.~Bai, G.~Wang, J.~Wang, X.~Yang, H.~Gao, X.~Liang, A.~Wang, M.~Islam, and H.~Ren, ``Ossar: Towards open-set surgical activity recognition in robot-assisted surgery,'' \emph{arXiv preprint arXiv:2402.06985}, 2024.

\bibitem{xu2021artificial}
Y.~Xu, X.~Liu, X.~Cao, C.~Huang, E.~Liu, S.~Qian, X.~Liu, Y.~Wu, F.~Dong, C.-W. Qiu \emph{et~al.}, ``Artificial intelligence: A powerful paradigm for scientific research,'' \emph{The Innovation}, vol.~2, no.~4, 2021.

\bibitem{haque2021multimix}
A.~Haque, A.~Wang, D.~Terzopoulos \emph{et~al.}, ``Multimix: sparingly-supervised, extreme multitask learning from medical images,'' in \emph{2021 IEEE 18th International Symposium on Biomedical Imaging (ISBI)}.\hskip 1em plus 0.5em minus 0.4em\relax IEEE, 2021, pp. 693--696.

\bibitem{wang2023sam}
A.~Wang, M.~Islam, M.~Xu, Y.~Zhang, and H.~Ren, ``Sam meets robotic surgery: an empirical study on generalization, robustness and adaptation,'' in \emph{International Conference on Medical Image Computing and Computer-Assisted Intervention}.\hskip 1em plus 0.5em minus 0.4em\relax Springer, 2023, pp. 234--244.

\bibitem{cui2024surgical}
B.~Cui, M.~Islam, L.~Bai, and H.~Ren, ``Surgical-dino: adapter learning of foundation models for depth estimation in endoscopic surgery,'' \emph{International Journal of Computer Assisted Radiology and Surgery}, pp. 1--8, 2024.

\bibitem{oquab2023dinov2}
M.~Oquab, T.~Darcet, T.~Moutakanni, H.~Vo, M.~Szafraniec, V.~Khalidov, P.~Fernandez, D.~Haziza, F.~Massa, A.~El-Nouby \emph{et~al.}, ``Dinov2: Learning robust visual features without supervision,'' \emph{arXiv preprint arXiv:2304.07193}, 2023.

\bibitem{hu2021LoRA}
E.~J. Hu, Y.~Shen, P.~Wallis, Z.~Allen-Zhu, Y.~Li, S.~Wang, L.~Wang, and W.~Chen, ``Lora: Low-rank adaptation of large language models,'' \emph{arXiv preprint arXiv:2106.09685}, 2021.

\bibitem{ma2018modeling}
J.~Ma, Z.~Zhao, X.~Yi, J.~Chen, L.~Hong, and E.~H. Chi, ``Modeling task relationships in multi-task learning with multi-gate mixture-of-experts,'' in \emph{Proceedings of the 24th ACM SIGKDD international conference on knowledge discovery \& data mining}, 2018, pp. 1930--1939.

\bibitem{lin2024smooth}
X.~Lin, X.~Zhang, Z.~Yang, F.~Liu, Z.~Wang, and Q.~Zhang, ``Smooth tchebycheff scalarization for multi-objective optimization,'' \emph{arXiv preprint arXiv:2402.19078}, 2024.

\bibitem{caruana1993multitask}
R.~Caruana, ``Multitask learning: A knowledge-based source of inductive bias1,'' in \emph{Proceedings of the Tenth International Conference on Machine Learning}.\hskip 1em plus 0.5em minus 0.4em\relax Citeseer, 1993, pp. 41--48.

\bibitem{duong2015low}
L.~Duong, T.~Cohn, S.~Bird, and P.~Cook, ``Low resource dependency parsing: Cross-lingual parameter sharing in a neural network parser,'' in \emph{Proceedings of the 53rd annual meeting of the Association for Computational Linguistics and the 7th international joint conference on natural language processing (volume 2: short papers)}, 2015, pp. 845--850.

\bibitem{long2017learning}
M.~Long, Z.~Cao, J.~Wang, and P.~S. Yu, ``Learning multiple tasks with multilinear relationship networks,'' \emph{Advances in neural information processing systems}, vol.~30, 2017.

\bibitem{lu2017fully}
Y.~Lu, A.~Kumar, S.~Zhai, Y.~Cheng, T.~Javidi, and R.~Feris, ``Fully-adaptive feature sharing in multi-task networks with applications in person attribute classification,'' in \emph{Proceedings of the IEEE conference on computer vision and pattern recognition}, 2017, pp. 5334--5343.

\bibitem{kendall2018multi}
A.~Kendall, Y.~Gal, and R.~Cipolla, ``Multi-task learning using uncertainty to weigh losses for scene geometry and semantics,'' in \emph{Proceedings of the IEEE conference on computer vision and pattern recognition}, 2018, pp. 7482--7491.

\bibitem{chen2018gradnorm}
Z.~Chen, V.~Badrinarayanan, C.-Y. Lee, and A.~Rabinovich, ``Gradnorm: Gradient normalization for adaptive loss balancing in deep multitask networks,'' in \emph{International conference on machine learning}.\hskip 1em plus 0.5em minus 0.4em\relax PMLR, 2018, pp. 794--803.

\bibitem{sener2018multi}
O.~Sener and V.~Koltun, ``Multi-task learning as multi-objective optimization,'' \emph{Advances in neural information processing systems}, vol.~31, 2018.

\bibitem{misra2016cross}
I.~Misra, A.~Shrivastava, A.~Gupta, and M.~Hebert, ``Cross-stitch networks for multi-task learning,'' in \emph{Proceedings of the IEEE conference on computer vision and pattern recognition}, 2016, pp. 3994--4003.

\bibitem{liu2019end}
S.~Liu, E.~Johns, and A.~J. Davison, ``End-to-end multi-task learning with attention,'' in \emph{Proceedings of the IEEE/CVF conference on computer vision and pattern recognition}, 2019, pp. 1871--1880.

\bibitem{wallingford2022task}
M.~Wallingford, H.~Li, A.~Achille, A.~Ravichandran, C.~Fowlkes, R.~Bhotika, and S.~Soatto, ``Task adaptive parameter sharing for multi-task learning,'' in \emph{Proceedings of the IEEE/CVF Conference on Computer Vision and Pattern Recognition}, 2022, pp. 7561--7570.

\bibitem{shazeer2017outrageously}
N.~Shazeer, A.~Mirhoseini, K.~Maziarz, A.~Davis, Q.~Le, G.~Hinton, and J.~Dean, ``Outrageously large neural networks: The sparsely-gated mixture-of-experts layer,'' \emph{arXiv preprint arXiv:1701.06538}, 2017.

\bibitem{fan2022m3vit}
Z.~Fan, R.~Sarkar, Z.~Jiang, T.~Chen, K.~Zou, Y.~Cheng, C.~Hao, Z.~Wang \emph{et~al.}, ``M$^3$vit: Mixture-of-experts vision transformer for efficient multi-task learning with model-accelerator co-design,'' \emph{Advances in Neural Information Processing Systems}, vol.~35, pp. 28\,441--28\,457, 2022.

\bibitem{aoki2022heterogeneous}
R.~Aoki, F.~Tung, and G.~L. Oliveira, ``Heterogeneous multi-task learning with expert diversity,'' \emph{IEEE/ACM Transactions on Computational Biology and Bioinformatics}, vol.~19, no.~6, pp. 3093--3102, 2022.

\bibitem{hu2021unit}
R.~Hu and A.~Singh, ``Unit: Multimodal multitask learning with a unified transformer,'' in \emph{Proceedings of the IEEE/CVF international conference on computer vision}, 2021, pp. 1439--1449.

\bibitem{ye2022taskprompter}
H.~Ye and D.~Xu, ``Taskprompter: Spatial-channel multi-task prompting for dense scene understanding,'' in \emph{The Eleventh International Conference on Learning Representations}, 2022.

\bibitem{wu2022orthogonal}
S.-H. Wu, Z.-H. Zhan, K.~C. Tan, and J.~Zhang, ``Orthogonal transfer for multitask optimization,'' \emph{IEEE Transactions on Evolutionary Computation}, vol.~27, no.~1, pp. 185--200, 2022.

\bibitem{zhu2020lymph}
Z.~Zhu, D.~Jin, K.~Yan, T.-Y. Ho, X.~Ye, D.~Guo, C.-H. Chao, J.~Xiao, A.~Yuille, and L.~Lu, ``Lymph node gross tumor volume detection and segmentation via distance-based gating using 3d ct/pet imaging in radiotherapy,'' in \emph{International Conference on Medical Image Computing and Computer-Assisted Intervention}.\hskip 1em plus 0.5em minus 0.4em\relax Springer, 2020, pp. 753--762.

\bibitem{grimwood2020assisted}
A.~Grimwood, H.~McNair, Y.~Hu, E.~Bonmati, D.~Barratt, and E.~J. Harris, ``Assisted probe positioning for ultrasound guided radiotherapy using image sequence classification,'' in \emph{International Conference on Medical Image Computing and Computer-Assisted Intervention}.\hskip 1em plus 0.5em minus 0.4em\relax Springer, 2020, pp. 544--552.

\bibitem{ramesh2021multi}
S.~Ramesh, D.~Dall’Alba, C.~Gonzalez, T.~Yu, P.~Mascagni, D.~Mutter, J.~Marescaux, P.~Fiorini, and N.~Padoy, ``Multi-task temporal convolutional networks for joint recognition of surgical phases and steps in gastric bypass procedures,'' \emph{International journal of computer assisted radiology and surgery}, vol.~16, pp. 1111--1119, 2021.

\bibitem{du2020multi}
B.~Du, J.~Liao, B.~Turkbey, and P.~Yan, ``Multi-task learning for registering images with large deformation,'' \emph{IEEE journal of biomedical and health informatics}, vol.~25, no.~5, pp. 1624--1633, 2020.

\bibitem{xu2018less}
Z.~Xu, Y.~Huo, J.~Park, B.~Landman, A.~Milkowski, S.~Grbic, and S.~Zhou, ``Less is more: Simultaneous view classification and landmark detection for abdominal ultrasound images,'' in \emph{Medical Image Computing and Computer Assisted Intervention--MICCAI 2018: 21st International Conference, Granada, Spain, September 16-20, 2018, Proceedings, Part II 11}.\hskip 1em plus 0.5em minus 0.4em\relax Springer, 2018, pp. 711--719.

\bibitem{liu2020prediction}
Q.-P. Liu, X.~Xu, F.-P. Zhu, Y.-D. Zhang, and X.-S. Liu, ``Prediction of prognostic risk factors in hepatocellular carcinoma with transarterial chemoembolization using multi-modal multi-task deep learning,'' \emph{EClinicalMedicine}, vol.~23, 2020.

\bibitem{yao2021deepprognosis}
J.~Yao, Y.~Shi, K.~Cao, L.~Lu, J.~Lu, Q.~Song, G.~Jin, J.~Xiao, Y.~Hou, and L.~Zhang, ``Deepprognosis: Preoperative prediction of pancreatic cancer survival and surgical margin via comprehensive understanding of dynamic contrast-enhanced ct imaging and tumor-vascular contact parsing,'' \emph{Medical image analysis}, vol.~73, p. 102150, 2021.

\bibitem{seenivasan2023task}
L.~Seenivasan, M.~Islam, M.~Xu, C.~M. Lim, and H.~Ren, ``Task-aware asynchronous multi-task model with class incremental contrastive learning for surgical scene understanding,'' \emph{International Journal of Computer Assisted Radiology and Surgery}, vol.~18, no.~5, pp. 921--928, 2023.

\bibitem{zhang20213d}
Y.~Zhang, H.~Li, J.~Du, J.~Qin, T.~Wang, Y.~Chen, B.~Liu, W.~Gao, G.~Ma, and B.~Lei, ``3d multi-attention guided multi-task learning network for automatic gastric tumor segmentation and lymph node classification,'' \emph{IEEE transactions on medical imaging}, vol.~40, no.~6, pp. 1618--1631, 2021.

\bibitem{song2020end}
L.~Song, J.~Lin, Z.~J. Wang, and H.~Wang, ``An end-to-end multi-task deep learning framework for skin lesion analysis,'' \emph{IEEE journal of biomedical and health informatics}, vol.~24, no.~10, pp. 2912--2921, 2020.

\bibitem{kenton2019bert}
J.~D. M.-W.~C. Kenton and L.~K. Toutanova, ``Bert: Pre-training of deep bidirectional transformers for language understanding,'' in \emph{Proceedings of naacL-HLT}, vol.~1.\hskip 1em plus 0.5em minus 0.4em\relax Minneapolis, Minnesota, 2019, p.~2.

\bibitem{radford2019language}
A.~Radford, J.~Wu, R.~Child, D.~Luan, D.~Amodei, I.~Sutskever \emph{et~al.}, ``Language models are unsupervised multitask learners,'' \emph{OpenAI blog}, vol.~1, no.~8, p.~9, 2019.

\bibitem{kirillov2023segment}
A.~Kirillov, E.~Mintun, N.~Ravi, H.~Mao, C.~Rolland, L.~Gustafson, T.~Xiao, S.~Whitehead, A.~C. Berg, W.-Y. Lo \emph{et~al.}, ``Segment anything,'' in \emph{Proceedings of the IEEE/CVF International Conference on Computer Vision}, 2023, pp. 4015--4026.

\bibitem{ravi2024sam}
N.~Ravi, V.~Gabeur, Y.-T. Hu, R.~Hu, C.~Ryali, T.~Ma, H.~Khedr, R.~R{\"a}dle, C.~Rolland, L.~Gustafson \emph{et~al.}, ``Sam 2: Segment anything in images and videos,'' \emph{arXiv preprint arXiv:2408.00714}, 2024.

\bibitem{wu2024fedfmsl}
P.~Wu, K.~Li, T.~Wang, Y.~Dong, V.~C. Leung, and F.~Wang, ``Fedfmsl: Federated learning of foundations models with sparsely activated lora,'' \emph{IEEE Transactions on Mobile Computing}, 2024.

\bibitem{wang2024surgical}
G.~Wang, L.~Bai, W.~J. Nah, J.~Wang, Z.~Zhang, Z.~Chen, J.~Wu, M.~Islam, H.~Liu, and H.~Ren, ``Surgical-lvlm: Learning to adapt large vision-language model for grounded visual question answering in robotic surgery,'' \emph{arXiv preprint arXiv:2405.10948}, 2024.

\bibitem{chen2024robust}
S.~Chen, Y.~Ju, H.~Dalal, Z.~Zhu, and A.~Khisti, ``Robust federated finetuning of foundation models via alternating minimization of lora,'' \emph{arXiv preprint arXiv:2409.02346}, 2024.

\bibitem{selvam2024rapid}
K.~P. Selvam, R.~Ramos-Pollan, and F.~Kalaitzis, ``Rapid adaptation of earth observation foundation models for segmentation,'' \emph{arXiv preprint arXiv:2409.09907}, 2024.

\bibitem{paischer2024one}
F.~Paischer, L.~Hauzenberger, T.~Schmied, B.~Alkin, M.~P. Deisenroth, and S.~Hochreiter, ``One initialization to rule them all: Fine-tuning via explained variance adaptation,'' \emph{arXiv preprint arXiv:2410.07170}, 2024.

\bibitem{chen2017deeplab}
L.-C. Chen, G.~Papandreou, I.~Kokkinos, K.~Murphy, and A.~L. Yuille, ``Deeplab: Semantic image segmentation with deep convolutional nets, atrous convolution, and fully connected crfs,'' \emph{IEEE transactions on pattern analysis and machine intelligence}, vol.~40, no.~4, pp. 834--848, 2017.

\bibitem{raffel2020exploring}
C.~Raffel, N.~Shazeer, A.~Roberts, K.~Lee, S.~Narang, M.~Matena, Y.~Zhou, W.~Li, and P.~J. Liu, ``Exploring the limits of transfer learning with a unified text-to-text transformer,'' \emph{Journal of machine learning research}, vol.~21, no. 140, pp. 1--67, 2020.

\bibitem{mehta2021mobilevit}
S.~Mehta and M.~Rastegari, ``Mobilevit: light-weight, general-purpose, and mobile-friendly vision transformer,'' \emph{arXiv preprint arXiv:2110.02178}, 2021.

\bibitem{ouyang2023efficient}
D.~Ouyang, S.~He, G.~Zhang, M.~Luo, H.~Guo, J.~Zhan, and Z.~Huang, ``Efficient multi-scale attention module with cross-spatial learning,'' in \emph{ICASSP 2023-2023 IEEE International Conference on Acoustics, Speech and Signal Processing (ICASSP)}.\hskip 1em plus 0.5em minus 0.4em\relax IEEE, 2023, pp. 1--5.

\bibitem{zhang2023customized}
K.~Zhang and D.~Liu, ``Customized segment anything model for medical image segmentation,'' \emph{arXiv preprint arXiv:2304.13785}, 2023.

\bibitem{zhou2024exploring}
Y.~Zhou, Z.~Zhao, H.~Li, S.~Du, J.~Yao, Y.~Zhang, and Y.~Wang, ``Exploring training on heterogeneous data with mixture of low-rank adapters,'' \emph{arXiv preprint arXiv:2406.09679}, 2024.

\bibitem{gao2023transendoscopic}
H.~Gao, X.~Yang, X.~Xiao, X.~Zhu, T.~Zhang, C.~Hou, H.~Liu, M.~Q.-H. Meng, L.~Sun, X.~Zuo \emph{et~al.}, ``Transendoscopic flexible parallel continuum robotic mechanism for bimanual endoscopic submucosal dissection,'' \emph{The International Journal of Robotics Research}, p. 02783649231209338, 2023.

\bibitem{yang2023novel}
X.~Yang, H.~Gao, S.~Fu, R.~Ji, C.~Hou, H.~Liu, N.~Luan, H.~Ren, L.~Sun, J.~Yang \emph{et~al.}, ``A novel miniature transendoscopic telerobotic system for endoscopic submucosal dissection,'' \emph{Gastrointestinal Endoscopy}, 2023.

\bibitem{bai2025surgical}
L.~Bai, G.~Wang, M.~Islam, L.~Seenivasan, A.~Wang, and H.~Ren, ``Surgical-vqla++: Adversarial contrastive learning for calibrated robust visual question-localized answering in robotic surgery,'' \emph{Information Fusion}, vol. 113, p. 102602, 2025.

\bibitem{allan2019endovis17}
M.~Allan, A.~Shvets, T.~Kurmann, Z.~Zhang, R.~Duggal, Y.-H. Su, N.~Rieke, I.~Laina, N.~Kalavakonda \emph{et~al.}, ``2017 robotic instrument segmentation challenge,'' \emph{arXiv preprint arXiv:1902.06426}, 2019.

\bibitem{lin2022libmtl}
B.~Lin and Y.~Zhang, ``Libmtl: A python library for multi-task learning,'' \emph{arXiv preprint arXiv:2203.14338}, 2022.

\bibitem{he2016deep}
K.~He, X.~Zhang, S.~Ren, and J.~Sun, ``Deep residual learning for image recognition,'' in \emph{Proceedings of the IEEE conference on computer vision and pattern recognition}, 2016, pp. 770--778.

\bibitem{hazimeh2021dselect}
H.~Hazimeh, Z.~Zhao, A.~Chowdhery, M.~Sathiamoorthy, Y.~Chen, R.~Mazumder, L.~Hong, and E.~Chi, ``Dselect-k: Differentiable selection in the mixture of experts with applications to multi-task learning,'' \emph{Advances in Neural Information Processing Systems}, vol.~34, pp. 29\,335--29\,347, 2021.

\bibitem{fernando2023mitigating}
H.~Fernando, H.~Shen, M.~Liu, S.~Chaudhury, K.~Murugesan, and T.~Chen, ``Mitigating gradient bias in multi-objective learning: A provably convergent approach.''\hskip 1em plus 0.5em minus 0.4em\relax International Conference on Learning Representations, 2023.

\bibitem{senushkin2023independent}
D.~Senushkin, N.~Patakin, A.~Kuznetsov, and A.~Konushin, ``Independent component alignment for multi-task learning,'' in \emph{Proceedings of the IEEE/CVF Conference on Computer Vision and Pattern Recognition}, 2023, pp. 20\,083--20\,093.

\bibitem{lin2023dual}
B.~Lin, W.~Jiang, F.~Ye, Y.~Zhang, P.~Chen, Y.-C. Chen, S.~Liu, and J.~Kwok, ``Dual-balancing for multi-task learning,'' 2023.

\bibitem{chen2017rethinking}
L.-C. Chen, G.~Papandreou, F.~Schroff, and H.~Adam, ``Rethinking atrous convolution for semantic image segmentation,'' \emph{arXiv preprint arXiv:1706.05587}, 2017.

\end{thebibliography}
\end{document}